# It Takes Two to Negotiate: Modeling Social Exchange in Online Multiplayer Games

KOKIL JAIDKA, National University of Singapore, Singapore
HANSIN AHUJA, Indian Institute of Technology Ropar, India
LYNNETTE NG, Carnegie Mellon University, USA

Online games are dynamic environments where players interact with each other, which offers a rich setting for understanding how players negotiate their way through the game to an ultimate victory. This work studies online player interactions during the turn-based strategy game, Diplomacy. We annotated a dataset of over 10,000 chat messages for different negotiation strategies and empirically examined their importance in predicting long- and short-term game outcomes. Although negotiation strategies can be predicted reasonably accurately through the linguistic modeling of the chat messages, more is needed for predicting short-term outcomes such as trustworthiness. On the other hand, they are essential in graph-aware reinforcement learning approaches to predict long-term outcomes, such as a player's success, based on their prior negotiation history. We close with a discussion of the implications and impact of our work. The dataset is available at https://github.com/kj2013/claff-diplomacy.

Online games such as Among Us and League of Legends are popular worldwide, with millions of daily users. These games are dynamic and interactive online environments where players can participate simultaneously over the Internet and communicate with their friends and strangers over voice, text, and video.

In-game communication offers an opportunity for understanding social exchanges, where speakers engage in dialogue to negotiate *mutual benefits* from the exchange. A mutual benefit, in the context of games, means that each participating player gains an advantage relative to their maximin benchmark [67]. The Social Exchange Theory (SET) discusses how relationships are built through exchanges that help people relate to each other [38, 68]. In this respect, while several studies explore the persuasive strategies employed in altruistic requests [2, 14, 74, 77], and sales pitches [36], besides some exceptions [36], they have rarely been studied in situations involving mutual benefit. The lack of research on the structure of online negotiations is a crucial research gap that has resulted in a limited understanding of modern digital culture. Beyond the gaming context, understanding the structure of online exchanges is critical to the field of computer-mediated communication, including but not limited to computer-supported cooperative work.

In this paper, we examine the research gaps that remain in understanding the broader negotiation strategies at work. While some qualitative work has been done on in-game communication, such as how players share information about gameplay [30], discuss potential risks and failures [43] and negotiate trades of mutual benefit [15], a comprehensive characterization of these strategies is lacking. Another research gap lies in understanding the long-term effects of the linguistic features of these exchanges. Some studies have examined the immediate impact of communication strategies on gameplay. For instance, trust is one of the many socio-emotional outcomes of these online interactions. The truster displays their confidence in the future actions of the trustee while undertaking a personal risk, despite the availability of other less risky actions [56]. However, prior work offers a limited understanding of in-game trust and betrayal [exceptions are 52, 60], and none consider the long-term effects of the linguistic features of these exchanges, offering another opportunity for further research.

This paper delves into the characteristics of negotiation in online interactions between players of a board game, Diplomacy, as well as their association with player trust and player success. Diplomacy is a seven-player game set on a pre-World War 1 European map, comprising 72 regions and 34 supply centers. Players represent European powers (England, France, Germany, Italy, Austria, Russia, Turkey) and control their home centers. The game has four phases yearly: two for communication and order finalization and two for order execution. Orders can direct units to stay, move, or support, with specifics like moving across land or water [4]. While players may agree to a draw, the game ends when a player controls 18 supply centers. While board positions are public, orders are secret, making it a game of imperfect information [61]. In recent research, Artificial Intelligence (AI) agents have been developed to play Diplomacy [3, 45]. For instance, a recent project employed planning and reinforcement learning algorithms that inferred players' beliefs and intentions from their conversations, together with a language model to negotiate and coordinate with other players. The agent surpassed human performance in 40 games of online Diplomacy with real participants, oblivious to the AI agent playing in their midst [3]. The study by Kramár et al. [45] extrapolates negotiating AI agents beyond a single participant to explore the problems in agent coordination and honesty, offering the insight that penalizing defecting peers could help improve in-game cooperation. Both these papers offer vastly different approaches to the problem of negotiation in Diplomacy, with the former focusing on simple goals such as honesty or defection and helpfulness, while the latter focuses on the mathematically optimal solution for mutual benefit. However, in efforts to make these problems computationally tractable, these studies had to ignore the other rhetorical possibilities, such as sharing information, establishing a rapport, or stalling for time, and even the possibility of more than two players, which implies combinatorially large action spaces with stochastic agent policies. In summary, studying interpersonal messages exchanged during a game of Diplomacy offers exciting opportunities to study goal-oriented, open-ended communication, where alliances, betrayals, and information sharing are crucial.

Four main questions guide our research on player negotiations in gameplay:

- **RQ1**: What linguistic features characterize players' negotiation strategies in online games involving social exchange? This question lets us understand language patterns regarding player negotiation strategies during gameplay.
- **RQ2:** How effectively can a player's negotiation strategies predict their perceived trustworthiness? This question extends the initial work on the linguistic predictors of trust to understand the broad negotiation strategies that earn trust in gameplay.
- **RQ3:** How well can a player's historical negotiation strategies predict their ultimate success in the game? This question considers whether turn-level decisions predict players' eventual chances of winning the game.
- **RQ4:** How well can a player's social influence predict their success, over and above the historical information about their negotiation strategies? This question considers how players accrue social capital during a game and whether it affects their chances of winning.

Our findings shed light on the linguistic markers of negotiation strategies, the predictors of trust and betrayal, the impact of negotiation strategies on long-term success, and the role of social influence in a player's ultimate victory. Our contributions include the following:

- A new taxonomy of negotiation strategies for mutual benefit grounded in prior theoretical and empirical work
- A newly annotated game dataset[1] labeled for the presence of negotiation strategies
- Models to predict negotiation strategies and their combined effectiveness in predicting the long- and short-term outcomes of player interactions

Our research approach bridges the gaps between computational linguistics and game economics to build an enriched understanding of social dynamics in online environments. Our findings about the importance of information and influence in online gameplay offer valuable insights for game developers, researchers, and players.

## 1 RELATED WORK

Research on cooperation and social interaction in multiplayer online games has delved into the design features that promote social interaction and cooperative gameplay, such as game patterns [28], awareness cues [76], and text messaging [15]. Studies on social interactions in multiplayer games have mainly focused on highlighting the importance of cooperation and interdependence in enhancing the gaming experience and strengthening social bonds among players [22, 24]. For instance, the use of "pings" or non-verbal cues in games like League of Legends has been shown to subtly affect individual team performance [47]. Research has also delved into the role of repeated social interactions [22, 25, 73], suggesting that they foster a sense of connection and cooperation when players engage with each other. However, the accrued effects of textual exchanges on participants' lives, such as their perceptions and ultimate victory at a game, remain unexplored.

Computational argumentation, a field that identifies and represents argument structures in text [66], can provide insights into these negotiations. The study of conversations or exchanges between participants is a common aspect of computational argumentation. Several studies have explored the predictive ability of arguments in online conversations on their outcomes[39, 41, 52, 77]. For example, research has quantified persuasiveness in advocacy requests [77], studied signals in conversational dyads to predict relationship durability [52], and evaluated different artificial game-playing agents implementing negotiation strategies [41].

To study the characteristics and the importance of computational argumentation in online multiplayer games, we have mined features from dyadic textual exchanges in a multi-party deliberative setting using human annotations to identify argumentative strategies and social network analysis to quantify measures of social influence. The dataset of Diplomacy collected by Peskov et. al [60] provides a rich context for this exploration. Diplomacy is a political strategy game where each player is a European power aiming to conquer the map by placing one's armies and negotiating with other players. Previous studies have studied fine-grained linguistic markers and turn-based atomized outcomes in Diplomacy [52, 60]. While insightful, focusing on discourse features ignores the broader strategies underlying gameplay. In this paper, we build on prior work by characterizing and annotating the different persuasion strategies evinced in the text messages exchanged between players. We thereby pose our first research question:

- **RQ1:** What linguistic features are characteristic of the different negotiation strategies players use in online games involving social exchange?

[1] https://github.com/kj2013/claff-diplomacy

While there are many outcomes of social exchanges that are worthy of study [19], our research focuses on how they build or degrade trust perceptions and how they boost a player's chances at winning the game. First, with reference to trust perceptions, the work by Niculae et al. [52] identifies some linguistic harbingers of betrayal, such as sudden shifts in emotional valence or mentions of future moves. There is a need to connect these to the broader negotiation strategies that drive the use of discourse markers. By doing so, we can build theoretical contributions regarding the roles of information sharing, self-disclosure, and friendliness in productive social exchanges. Therefore, we raise the second research question:

- **RQ2:** How well do players' negotiation strategies predict their perceived trustworthiness?

Next, with reference to player success, we are interested in whether social exchanges can offer insights into long-term benefits, such as whether a player ultimately wins a game. A preliminary study of simulated online negotiations suggests a relationship between long-term outcome satisfaction and trust change [80]. Prior research lacks an exploration of the long-term economic benefits of these social exchanges, such as a player's ultimate victory after weeks of negotiating gameplay. Formulating a prediction about the winner of a strategy game should consider each player's strategy and the relationships they accrued, evidenced in their exchanges during the game. While prior research on creating Diplomacy bots has experimented with predicting player success based on the counter-offers accepted or refused in recent history [54], they use a "no-press" setting without the possibility for players to exchange chats. Ours is the first study to explore the predictive effect of text-based negotiation strategies on player victory, and we pose the following research question:

- **RQ3:** How well does historical information of a player's negotiation strategies predict their ultimate success?

Social exchanges in online environments are pivotal in fostering trust, promoting cooperation, and solidifying interpersonal relationships, thereby contributing to the accumulation of social capital [38, 68]. Previous research has underscored the correlation between online interactions and social capital, indicating that sustained interactions with other players can facilitate the accrual of social capital. For instance, a study by Bisberg et al. [6] emphasized the significance of social influence by examining the contagion of player generosity. Despite these insights, there remains a gap in understanding how the social influence gained through these exchanges impacts the long-term outcomes of online interactions. Incorporating social capital in models of player success may offer more accurate predictions and motivates our fourth research question:

- **RQ4:** How well does a player's social influence predict their success, over and above the historical information about their negotiation strategies?

## 2 METHOD

In this work, we followed a machine learning approach on a dataset of online players' exchanged messages to characterize the in-game negotiation dynamics for mutual benefit. We identified negotiation strategies in online chat messages and trained classical and neural network classifiers to predict these strategies and evaluate their effects on the perceived trustworthiness of a player. Finally, we have evaluated the efficacy of negotiation strategies and in-game facts (such as score-based power differentials) to predict player victory. This section details these steps of the method, followed by a description of the experimental setup.

Table 1. The taxonomy of negotiation strategies (in bold) for mutual benefit, their definitions, exemplifications (in italics), and connections with prior work.

| Category | Strategies (% Positive instances ) |
|---|---|
| **Ethos** [13, 72, 74]: The player establishes personal credentials through the sharing of information about their thoughts and moves, or moves by the recipient or other game players. | **Speaker's move (N = 5,918, 25.41% Positive Instances)**:<br>• *Plans:* I'm attempting to make that deal with Russia now (...)<br>• *Plans:* I'm actually sort of running counter-intelligence for England (...)<br>• *Thoughts:* I think me and England are really on the same page at this point regarding France<br>• *Goals:* And I am committed to supporting Munich holding.<br>**Recipient's move (N = 5,918, 44.69% Positive Instances)**:<br>• *Propose a plan of action:* Make sure you don't move Munich so that it can take my support.<br>• *Counter-offer:* Well, are you willing to humor my question about the Aegean, anyway?<br>• *Seek clarification:* Are you willing to tell me what your plans are for the Tri unit, or at least to warn me before any move into Tyrolia?<br>**Other player's move (N = 5,918, 9.58% Positive Instances)**:<br>• *Sharing information:*<br>– France held out a long time<br>– He has a serious Austria problem.<br>– That's not what Austria said to England |
| **Logos [18, 32]**: The sender anticipates future moves, offers justification or explanations for a move by themselves or by the receiver, or discusses a move that already happened. | **Reasoning (N = 11,032, 79.21% Positive Instances)**:<br>• *Speculation:*<br>– France is a really good player, and he is no doubt working hard to get England to turn on you.<br>– Why would France help us?<br>– I think that England will want to coax me to attack you with him after France falls<br>– He stopped talking to me, so I bet he's trying to turn England.<br>• *Justificaton:*<br>– If you took Marseilles, I would be stronger against England<br>– This will improve all of our chances of crushing France quickly.<br>– I like the unit there because it sets up an attack on Austria if I ever want to go that route (build A Ven and go east).<br>• *Hindsight:*<br>– I probably should have just told you my moves;<br>– You could have advised me that supporting Mun-Bur was more important than Kie-Ruh |
| **Pathos** [**17**, **77**]: The sender shows friendliness to the receiver either through sharing personal information or thoughts, general banter, reassurances, compliments, or apologies. | **Friendliness (N = 5,918, 65.49% Positive Instances)**:<br>• *Sharing personal information:* But in the interest of continued full disclosure, here's what I think (...)<br>• *General banter:* Nah, I just needed some reassurance :)<br>• *Reassurance:* You are my favorite.<br>• *Greet, thank, or compliment:* I didn't know that! Thanks!<br>• *Apology:* Ha! So sorry!! I meant that for France |

## 2.1 Negotiation Taxonomy

The motivation to explore the effect of different player strategies arose from the theoretical gaps in the role of interpersonal interaction in social exchange [for a discussion, see 19]. We are interested in negotiated exchanges, specifically on the different modes of persuasion and their effect on the outcomes of an interaction aimed at mutual benefit [21, 49]. Prior work by Carlile et al. [13], Hidey et al. [37] and others relate the three modes of persuasion – Ethos, Pathos, and Logos (defined in Table 1) – to their linguistic characteristics. Cialdini and Garde [17] characterized authority as a marker of Ethos, while Duthie et al. [26] developed a text analysis pipeline to mine Ethos from political debates, making use of named-entity resolution, parts-of-speech, and sentiment analysis. Habernal and Gurevych [32] annotated argument pairs for Logos, or logical reasoning through examples, facts or game statistics [18]. Studies of Pathos, or use of emotional aspects, in conversation exchanges often refer to likeability and reciprocity [17, 77]. Chen and Yang [14] extended the basic framework to identify persuasion strategies such as emotion, credibility, and impact in altruistic requests.

Through preliminary data annotation using the coding schemes provided in prior work, we observed that existing taxonomies mainly exemplified strategies from Pathos. We recognized the need to update the understanding of persuasion strategies as a response to others. Prior research has also reported on how multiplayer games require players to consider whether to follow a unique or a coherent strategy for team effectiveness [42], to cooperate with other players or compete to win the game [22, 62], and to mediate conflicts when they arise among co-players [64]. Therefore, we observed that Logos and Ethos appeared to play an important role in Diplomacy. For instance, players would share in-game information (*"I'm attempting to make that deal with Russia now"* ). Players would also speculate on future moves by other players (*"I think that England will want to coax me..."* ). Therefore, as our first contribution, we created an extended categorization of negotiation strategies that (a) applies prior work to paradigms for mutual benefit and (b) is more closely in the persuasive modes they reflect. We categorized some strategies as **Ethos** if the sender used them to establish personal credibility, e.g., through self-disclosure and information-sharing. **Logos**, on the other hand, comprised strategies where the sender offered justification or explanations for a move in the past or the future. Finally, strategies categorized as **Pathos** demonstrated friendliness or reassurance to the recipient. The detailed description of the labels is in Table 1. In the following paragraphs, we have discussed how we evaluated our negotiation taxonomy through inter-annotator reliability statistics.

## 2.2 Annotation procedure

We processed the Diplomacy dataset (N = 13,132) by Peskov et. al [60] by segmenting each chat message into sentences. This dataset consists of pairwise online chat conversations among players of the Diplomacy game. Each message was annotated (at the time of sending or receiving the message) as actual 'Truth' or 'Lie' and perceived 'Truth' or 'Lie' by the sender and the recipient, respectively. Then, we filtered out sentences that were less than five words to ensure the messages provided sufficient information to offer a reliable judgment. Finally, to control for individual differences and sender-side confounds, we filtered and sampled 16,000 sentences intended as 'Truths' by the sender. These formed the basis of the two-stage annotation process [29, 65, 70]. In either stage of the annotation, the annotators underwent a training task on a small subset before beginning work on the actual annotation.

In the first step, an Amazon Mechanical Turk task was used to crowdsource labels corresponding to different rhetorical strategies. The worker criteria comprised residents of the United States with a minimum approval rate of 80% and a minimum of 1000 accepted hits.[2] Each sentence was assigned to five annotators. Annotators annotated sentences for each of the four labels (part 1 of Table 2) in a binary format: 1 for the presence of the feature in the sentence and 0 for the absence. Labels are not mutually exclusive. For example, annotators could label a sentence with both friendliness and reasoning. Following the recommendation by Passonneau and Carpenter [55] and being cognizant of the limitations of chance-based agreement measures, we have also reported the probabilistic model-based inference of agreement besides the pairwise percentage agreement. Unlike chance-based metrics, which have wide error bounds, model-based measures consider the actual categories of items in the corpus and the prevalence of each label in the corpus to ultimately report annotators' accuracy by category, which we have reported as the average of the true-positive and true-negative accuracy under $\theta$ in Table 2. Based on recommended thresholds [55], we considered the label-level annotation quality acceptable for training machine learning models if it hit at least a pairwise percentage agreement of 75% and a $\theta >= 0.65$, which implied that the annotations for only one label qualified for training machine learning classifiers.

[2]The instructions provided to the annotators are provided in the supplement.

Table 2. Inter-annotator reliability statistics. $\theta$ reflects the average annotator accuracy across true-positives and negatives. The right-most column depicts whether the inter-annotator agreement was deemed sufficient to use the labels for model training, or whether the data was reannotated with expert annotators and a refined coding scheme. The label definitions are the same as those reported in Table 1.

| | AMT Annotation (N = 16,000; J = 5 annotators) | | | |
|---|---|---|---|---|
| | Pairwise % Agree | % with 80% agreement | $\theta$ (Average accuracy) | Used for model training |
| **Reasoning** | **75.26** | 57.26 | **0.66** | **YES** |
| **Speaker's or Recipient's move** | 81.47 | 73.49 | 0.63 | NO |
| **Other player's move** | 76.78 | 57.82 | 0.59 | NO |
| **Friendliness** | 76.54 | 60.97 | 0.58 | NO |
| | **Expert reannotation (N = 6000; J = 5 annotators)** | | | |
| | **Pairwise % Agree** | **% with 80% agreement** | **$\theta$ (Average accuracy)** | **Used for model training** |
| **Speaker's move** | 73.93 | 56.95 | 0.67 | YES |
| **Recipient's move** | 73.43 | 52.21 | 0.65 | YES |
| **Other player's move** | 89.67 | 85.12 | 0.74 | YES |
| **Friendliness** | 79.23 | 65.18 | 0.65 | YES |

Subsequently, we organized a second annotation task in which we (a) simplified the annotation instructions, (b) simplified the coding scheme for a finer-grained distinction between messages discussing game moves, and (c) employed and trained expert annotators. We also deduplicated our annotation dataset by grouping similar sentences (> 0.8 similarity) through a pairwise cosine similarity calculation between the word vectors of the sentences constructed using the BERT model, which reduced the dataset size by about 4000 sentences. Subsequently, for the second round of annotation, we had a smaller random stratified subset of 6,000 sentences. Five trained annotators with a Master's degree in Linguistics were first trained, and disagreements in a pilot task were resolved through discussion. Then, they re-annotated the full dataset independently. As a result, we obtained new labels for evidence of discussion of game moves, either about the speaker, the recipient, or the other player, with improved $\theta$ values ranging from 0.65-0.74.

## 2.3 Approach

We followed a two-step annotation process to create a labeled dataset of negotiation strategies that characterize textual conversation exchanges involving mutual benefit. Next, we reported the linguistic characteristics predictive of different negotiation strategies to address **RQ1**. Subsequently, to address **RQ2**, we evaluated the effectiveness of our taxonomy at predicting short-term outcomes, such as player trustworthiness. We included the negotiation labels in linear mixed models and a combined classifier trained on fine-tuned DistilBERT embeddings to predict message trustworthiness.

In the second part of our analysis, we evaluated the impact of a player's historic choice of strategies and their social influence across multiple dyadic interactions in the predictive performance of player victory. To address **RQ3**, we evaluated the effectiveness of persuasion strategies in predicting long-term outcomes. We used weakly supervised learning methods to generate labels on additional data and applied them in a reinforcement learning approach to predict player victory. Then, in experiments for **RQ4**, we further demonstrated the advantage of hybrid, graph-aware approaches that enrich negotiation models with social influence information. Finally, we reported the improvements offered by different operational choices on the predictive performance.

2.3.1 *Training classifiers on the negotiation labels.* In keeping with best practices for text classification setups that are reported elsewhere [20], only the labels with at least 75% agreement (which constituted 64.8% of all labels, individual N's are reported in Table 1) were subsequently used in training and testing neural network classifiers to label the strategies evinced at the message-level.

We evaluated fifteen approaches for text classification for each training set and ultimately chose the method with the highest performance. The fifteen approaches include nine classical supervised machine learning models (i.e., logistic regression, decision trees) and six neural network models (i.e., BERT-based, ROBERTA-based). The neural network models performed better and were used to construct the final model. For full results, refer to Table 7, Table 8 and Table 9. In the following paragraphs, we have described the setup of the neural network classifiers and reported the other classifiers in the Appendix.

**Neural Network Classifiers.** We used five neural network classification approaches provided by the Python library *SimpleTransformers*[3]: BERT, RoBERTa, ALBERT, DistilBERT, and XLNet. These models are constructed using the transformers architecture, which means every input element of one step is connected to every output element of the subsequent step. The weights of the input elements are dynamically calculated based on their connections. In addition, we also used the GPT-3 classifier [10].[4] We also include two classifiers as a baseline comparison for the machine learning approaches - the Random label selection and the Majority label classifier. The Majority label classifier assigns the most prevalent label in the dataset to all data points. The Random label classifier randomly assigns a label to each data point. For each classifier, we subdivided the dataset for a 10-fold cross-validation run using a stratified split, which maintains the proportion of each data class within the splits. We ran each model for eight epochs, after which the model loss is less than $\delta = 0.001$. For the other parameters, we used the default values provided by the library.

2.3.2 *Short-term consequences: Perceptions of trustworthiness.* We evaluated whether a player's chosen negotiation strategy predicts the message's trustworthiness. Each message in the Diplomacy dataset includes the label of *Trustworthiness*, indicating whether the recipient thought the sender was telling the truth in their message. A perceived truth is given a positive label, while a perceived lie is given a negative label. We formulated this analysis as linear mixed models with fixed random effects corresponding to the specific game being played and the duration of the conversation between a pair of speakers.

The dataset comprised the entire Diplomacy corpus. The model inputs comprised the predicted labels of negotiation strategies, generated using the best-performing classifiers on individual label prediction tasks, reported in Table 3.

2.3.3 *Long-term consequences: Winning the game.* We used graph-aware reinforcement learning to model a player's victory in the game. The technique consists of two steps: extracting each player's game state and action information at each chat message and then formulating the winning conversation thread as a score-based inverse reinforcement learning problem [27].

Each Diplomacy game is characterized by a conversation thread $t$,

$$t = (s_0, \ldots, s_T) = (s_j)_{j=0}^{T}$$

[3]https://github.com/ThilinaRajapakse/simpletransformers. Citations are reported in the Appendix
[4]Details about the approach followed by each model are provided in the Appendix.

Where each state $s$ of the state space S is encoded using $\phi : S \rightarrow R^d$. We developed three variants of the encoded state space. First, we operationalized an encoding that relies on the player's score at any given point, which we denote as "SBIRL". Second, we evaluated one that only incorporates the player's social influence by incorporating eight measures of the player's centrality in the game-specific social network. These features are elaborated on in the Experimental setup section, and we named this variant "Graph-only SBIRL." Finally, we tested a combination of the two, "Graph-aware SBIRL." For each player, we created thread-level tuples $(t_i, f_i)$, where $t_i$ is the subsequence of states corresponding to the $i^{th}$ player (referred to as a subthread) and $f_i$ is the final score of the $i^{th}$ player at the end of the thread.

Next, we calculated reward function $r$ as a discounted sum of rewards at the subthread level. Theoretically, $r_\theta(s) = \theta^\top\phi(s)$, where $\theta$ is the set of parameters and $\phi$ is a state in the state space. However, a discounting factor $\gamma$ is needed to re-calibrate the effect of states on the outcome.

Making appropriate substitutions, we expanded the operationalization of $r$ by including every discounted state within every subthread. We finally obtained:

$$\sum_{t=0} \gamma^t r_\theta(s_t) = \theta^\top\mu(h) \text{ with } \mu(h) = \sum_{t=0} \gamma^t \phi(s_t)$$

where $\gamma$ is the discounting factor.

Finally, we regressed the player's scores $f_i$ on the mappings $\mu(h_i)$ while asymptotically minimizing the risk using the $\ell_2$-loss. The eventual reward function estimator $r_{\theta_n}$ is derived through estimating $\theta_n$ with the equation:

$$\theta_n = \underset{\theta \in R^d}{\text{argmin}} \frac{1}{n} \sum_{i=1} (f_i - \theta^\top\mu(h_i))^2$$

## 3 EXPERIMENTAL SETUP

Many of the final training datasets for the different labels reported in Table 1 had a skewed label distribution as reported in the heading rows in Table 1. The following paragraphs describe the dataset's features, including the social network features which were used to enrich reinforcement learning approaches trained on player victories.

**Linguistic features**:
To obtain linguistic insights about the negotiation strategies, we extracted a variety of content and discourse features from the training data:

- Discursive features: These include stylistic and psycholinguistic features. Stylistic features comprise scores for politeness, harbingers of betrayal, and psycholinguistic features in writing and have been applied to model politeness and trustworthiness in text [20, 52]. Psycholinguistic features denote emotional, cognitive, and social processes exemplified in writing and comprise the categories in Linguistic Inquiry and Word Count library [59]. Their association with user behavior has been reported in prior work [52, 78].
- Content features: We included two types of content features. The first is the count-vectorizer [5], which constructs a sparse matrix representation of the frequencies of words in the dataset and represents each message as a vector of these frequencies. The second is the Term Frequency-Inverse Document Frequency (TFIDF) vectorizer, which converts phrases into a frequency distribution weighted by their uniqueness in the overall dataset.

Next, these features were used to evaluate the performance of classical classifiers for predicting different negotiation strategies before we experimented with neural network models that offered a substantive improvement boost. We evaluated the predictive performance of classical classification approaches trained on several combinations of features:

[5] https://scikit-learn.org/stable/modules/generated/sklearn.feature_extraction.text.CountVectorizer.html

(1) discursive (*discursive-features*); (2) content features via word count (*word-features*); (3) content features via TFIDF (*tfidf-features*); (4) combining content features via TF-IDF and discursive (*tfidf-discursive-features*). While the content-based approach was anticipated to have higher accuracy, the discursive approach would be more transferable across domains in a context-sensitive task. Examples of some of these categories, their definitions, and some of the underlying linguistic cues are provided in the Appendix.

**Social influence features**:
Social influence reflects how players adjust their interactions based on the other players. The reinforcement learning approach to predict player victory incorporated the social influence of the player accrued over time. We constructed a social graph of the player interactions, with players as the nodes, to implicitly capture the importance of players in the information network constituting the game. Edges reflect whether any messages were exchanged between two players, and edge weights denoted the frequency of such messages. Following the recommendation of other studies in using network measures to operationalize social capital [7], the following centrality measures were calculated and included in model training. Each feature reflects a different perspective of the sender's social influence, such as their importance as a coordinator, representative, gatekeeper, itinerant, or liaison [12, 31]. For instance, individuals with greater centrality can act as "brokers" with access to diverse information and knowledge pools. In this manner, they may be able to control how information flows between different social groups [7, 9, 12, 53]:

- **Eigenvector centrality**: It is proportional to the sum of the centralities of those recipients with which a sender is connected.
- **Closeness**: A measure of the degree to which a sender is near all other individuals in a network.
- **PageRank**: A measure of the number of times the sender is encountered in a random walk over the social network.
- **Subgraph density**: A measure of the number of edges to the number of vertices for the subgraph containing the sender.
- **Betweenness centrality**: A measure of the number of the shortest paths connecting nodes that pass through a particular node.
- **Hub and Authority score**: The Authority scores for a node in a network (and a Sender in a game) reflect the value of its content. The Hub score reflects the degree the node is linked to other nodes. The two measures distinguish nodes that are important but less connected to the other well-connected nodes.

### 3.1 RQ1: Interpretability analysis

In order to provide linguistic insights around the different negotiation strategies, we used the *SHAP* library[6] on the outputs from the best-performing classical classifiers. The *SHAP* (SHapley Additive exPlanation) library draws ideas from game theory, visualizing the Shapely value for each feature [50]. The Shapely value is the average marginal contribution of the feature among the possibilities of occurrences in the messages, indicating the impact of the feature on the prediction. Next, we used the *transformer-interpret* library[7] on the outputs of the best-performing neural network classifiers. The *transformers-interpret* library approximates the contributions of words to a neural network classifier prediction through integrated gradients computed for each input feature word.

[6]https://github.com/slundberg/shap
[7] https://github.com/cdpierse/transformers-interpret

### 3.2 RQ2: Model training and Weakly supervised labeling

The next step involved defining the action space for the players in the Diplomacy games in line with the expectation of a reinforcement learning approach. First, a weakly supervised approach was followed to predict the rhetorical strategies for the entire dataset. The training set comprising the labels from CL-Aff Diplomacy was used to train binary classifiers on the different rhetorical strategies, such as Friendship, Reasoning, Game Move, and Share Information. Next, the best-performing classifiers (Table 3) were used to predict labels at the message level for the entire dataset. Finally, the labels were included to answer RQ2 by (a) comparing the predictive effects of different strategies in linear mixed models and (b) training ensemble classifiers to jointly predict the Trustworthiness label, using class weights in the hyperparameter settings, as only 4.3% of cases were perceived as untrustworthy. The labels were also used to learn states to answer RQ3, as described in the following paragraphs.

### 3.3 RQ3-RQ4: Reinforcement learning setup

The reinforcement learning task required mutually exclusive action states for each participant at each message timestamp. Therefore, we devised a new binary label we call 'Action state,' which reflected the sender's action in each message under assumptions of mutual exclusion. First, pairwise correlations between player strategies were calculated to identify which strongly correlated with each other, on which basis two anti-correlated strategies seemed to emerge. At the player level, Group 1 comprised Reasoning, Game moves (a union of Speaker's move and Recipient's move), and Other player's moves, which all shared a strong pairwise Pearson correlation ($r \in (0.45, 0.55)$). At the same time, each was strongly anti-correlated with Group 2, comprising only Friendliness ($r \in (-0.61, -0.45)$), both with $p < 0.001$. Following these insights, we computed the union of each message's labels to derive its Action state under assumptions of mutual exclusion. A majority voting mechanism was used in case two states were equally represented.

Next, we evaluated the predictive performance of different models that attempt to identify the winner of a game based on input parameters about player states and actions. We offer a comprehensive evaluation of different models to understand better the importance of linguistic and social influence predictors on the chances of winning.

One out of seven players per game (14.3%) was labeled as the winner. First, in random-state SBIRL, we encoded only the difference between the game scores of the two players in our state feature vector, as these fluctuated widely and were not correlated with the actual game outcome. Next, we evaluated the simple SBIRL, which addresses **RQ3**, as it relied solely on negotiation strategies. Finally, graph-only and graph-aware SBIRL address **RQ4** and include multiple graph centrality features, such as authority score and eigenvector centrality, into the state representation.

## 4 RESULTS

### 4.1 Linguistic Insights

To answer RQ1, the beeswarm plots in Figure 1 use SHAP values to show the distribution of the impact of the most critical features in the best-performing logistic regression models, as determined in the descending order of the total SHAP value magnitudes over all samples. The color represents the feature value (red high, blue low). For example, in Figure 1a, that a *high use* of the 'You' in a message is *the least* likely to predict Ethos, specifically, a negotiation strategy involving sharing information about the sender's moves.

Figure 1a also offers face validity as we observe that when discussing their moves, speakers are more likely to use first-person pronouns such as 'I.' They are also characteristically longer than average, with a higher word count. Prior research has discussed pronouns as an important indicator of the truthfulness or deception [58]. Similarly, Figure 1b suggests that when senders invoke a higher word count, it could also be predictive of discussing the receiver's move. They may also invoke a higher use of cognitive words such as 'should, would' that highlight discrepancies or words denoting time, such as 'end, until, season.' On the other hand, the higher use of words that appear to shift the focus onto other topics, such as first-person pronouns or impersonal pronouns, such as 'it's,' predicts the absence of this strategy.

Table 3. The predictive performance of the best-performing neural network machine learning classifiers on held-out data in a five-fold cross-validation setup. The full set of results are reported in the Appendix.

| | **Best-performing neural network classifiers** | | | |
|---|---|---|---|---|
| **Label** | **Approach** | **Accuracy** | **Macro F1** | **Minority F1** |
| **Other player's move** | ALBERT | 0.933 | 0.803 | 0.644 |
| **Speaker's Move** | RoBERTa | 0.834 | 0.783 | 0.678 |
| **Recipient's Move** | RoBERTa | 0.801 | 0.799 | 0.781 |
| **Reasoning** | DistilBERT | 0.712 | 0.514 | 0.204 |
| **Friendliness** | ALBERT | 0.760 | 0.738 | 0.663 |

In Figure 1c, we can observe that among all the inputs to the model, high use of six-letter words, verbs, and a present focus predicted a higher probability of an Ethos strategy involving other players' moves, on average. On the other hand, messages that had higher values of second-person pronouns, such as 'you,' started with the names of other players, or used more filler words, such as articles, predicted the absence of such a strategy. This is consistent with previous studies where the use of pronouns and type of pronouns are predictors of the presence of the Ethos strategy in persuasive short texts from social media exchanges [16, 71].

Figure 1d indicates that the use of Logos, specifically reasoning, as a negotiation strategy is predicted by the high use of words denoting anger, such as 'hate, kill, annoyed,' and cause, such as 'because,' 'effect.' Adverbs and connective phrases are also commonly used to connect phrases of reasoning [16]. On the other hand, the higher use of words denoting time predicts the absence of reasoning as a strategy.

Finally, Figure 1e indicates that higher use of pronouns, articles, informal words, and adjectives predicts Pathos or friendliness. The presence of a vast vocabulary and descriptive words within the texts predicts Pathos as users share evidence through explicit references or narratives to establish their credibility. On the other hand, the higher use of question marks predicts the absence of friendliness. Our findings corroborate prior work demonstrating the use of credibility as a persuasion strategy by Chen and Xiao [16] in the misinformation domain.

### 4.2 Internal validation

Table 3 reports the best-performing neural network classifiers in the internal validation.[8] The ALBERT models for other players' moves and friendliness and the DistilBERT model for reasoning, with fewer parameters and faster training time than the other models, also performed better in held-out validation. On the other hand, the large batch size and fewer training steps afforded by RoBERTa ensured its superiority in the models trained on the speaker's and recipient's moves.

Figure 2 reports the messages where the best-performing classifier correctly predicts the presence of a negotiation strategy, allowing us to observe the linguistic cues that enable the prediction. Transformers appear to have inferred the importance of first-person pronouns in mentioning the sender's moves, second-person pronouns in mentioning the receiver's moves, and third-person pronouns and country names in mentioning other players' moves. The models also appear to pay attention to question marks and discrepancy words ('if') in the recipient's moves, and verbs ('fighting,' 'evacuate'), adverbs ('actually'), and adjectives ('useful') that predict reasoning as a strategy. Finally, for friendliness, the model paid attention to words referring to relationships and honesty. The actual feature-prediction association may be more complex due to interactions between features, and these visualizations are mainly illustrative.

[8]The full results are reported in the supplement.

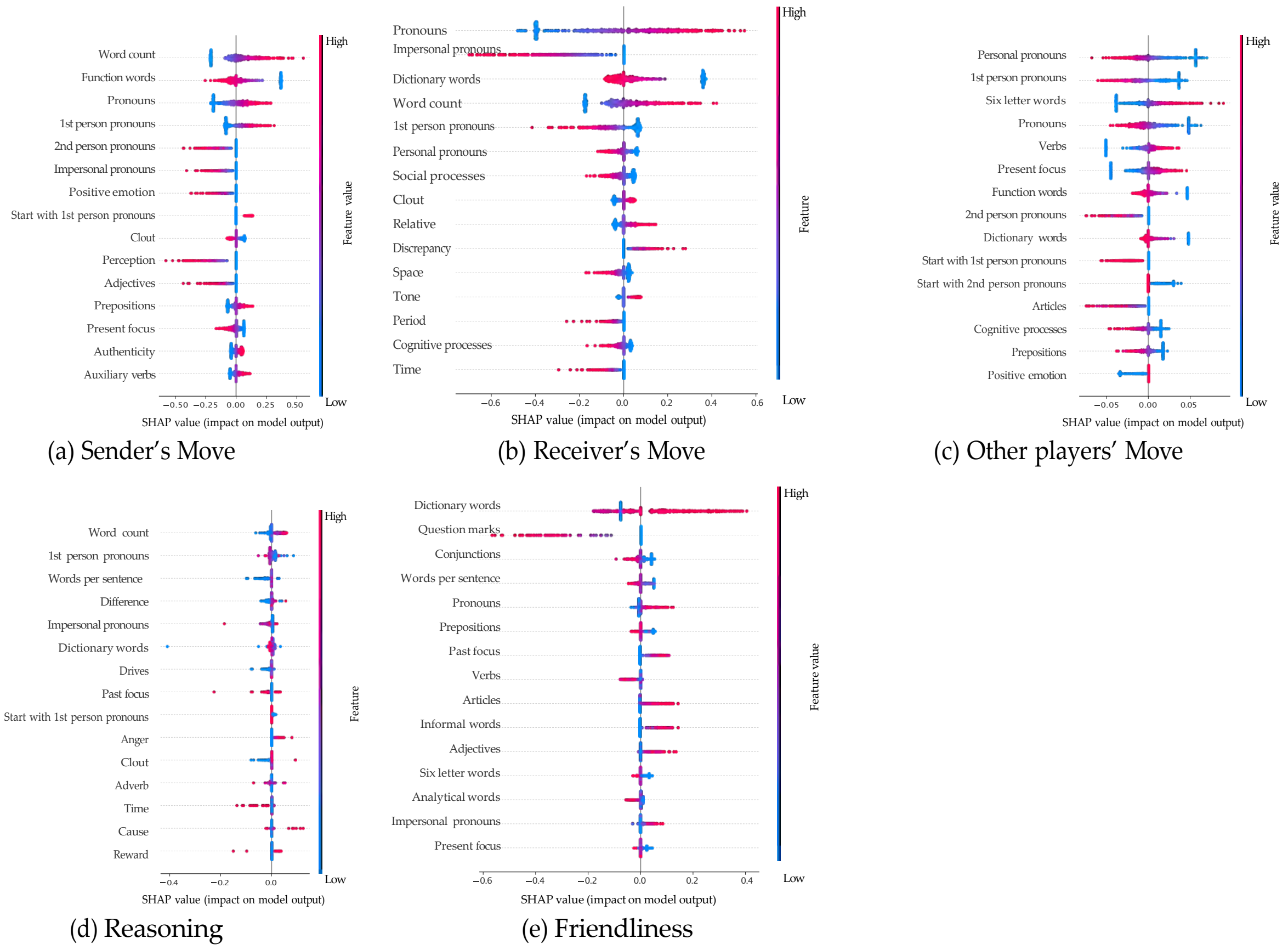

Fig. 1. Shapley value plots denoting the importance of different linguistic features in the classical classifiers. Best seen in color. The more positive the SHAP value range, the better positive predictive value the linguistic feature has for the corresponding negotiation strategy.

### 4.3 Short-term consequences: Perceptions of trustworthiness

We applied linear mixed models to predict trustworthiness as a function of negotiation strategies, with fixed effects that control for game-level differences and the duration of the conversation. The coefficient plot reported in Figure 3 reports the findings and allows a comparison of the predictive effects of negotiation strategies. We observe that, perhaps unexpectedly, self-disclosure and friendliness were significant *negative* predictors of trustworthiness. We will discuss the possible reasons behind this finding in the Discussion section.

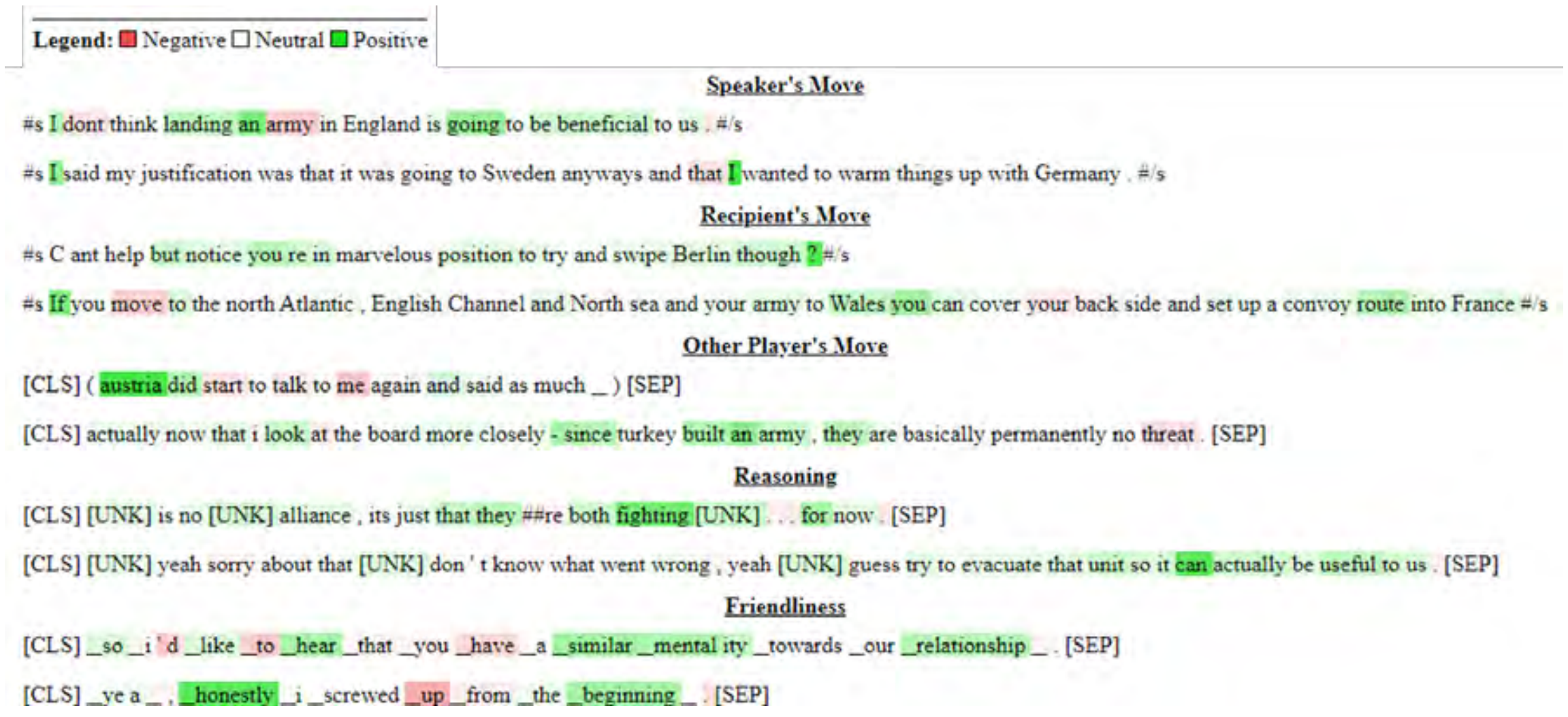

Fig. 2. Word attributions for positive classification for each negotiation strategy. Words highlighted in green (red) positively (negatively) attributed to the outcome; the darker the highlight, the higher the attribution. Best seen in color.

To estimate whether the strategies can jointly predict the perceived trustworthiness of a message, we also explored (a) directly predicting Trustworthiness through models trained on different linguistic features and (b) predicting Trustworthiness through a combined classifier approach, both with a ten-fold cross-validation setup. However, as reported in the first column in Table 4, we found that directly training trustworthiness models on language gave a poor performance. It is also important to note that the low overall F1 scores for predicting the perceived trustworthiness of a message corroborate those reported before us on the same data [60].

On the other hand, adding negotiation labels enriched the predictive performance. Similar approaches have been proposed in hierarchical classification models elsewhere [77]. However, combined approaches do not clarify the contributory effect of each strategy. The ablation analysis reported in Table 4 confirms Ethos's importance in training combined classifiers for Trustworthiness. We can observe that the most substantial contribution to the Minority-F1 appears to be through Ethos (although the minority-F1 numbers remain poor in general). As compared to the original paper [60], our evaluation is reported on a deduplicated subsample that is 45% in size (N~13,000 vs. N~6,000); therefore, our findings may not be directly comparable with the original authors. Nevertheless, we achieve F1 scores within 3% of their best score.

### 4.4 Long-term consequences: Winning the game

The winner of a chat thread is the player with the higher score at the end of the thread. The accuracy of the reward function is the fraction of times when the average estimated reward for the winner was greater than that of the loser. The results are reported in Figure 4a. The best-performing approach is the graph-aware SBIRL, returning an accuracy of 0.790, followed by graph-only SBIRL with 0.640 accuracy. The vast difference illustrates how a player's social influence predicts their success in Diplomacy.

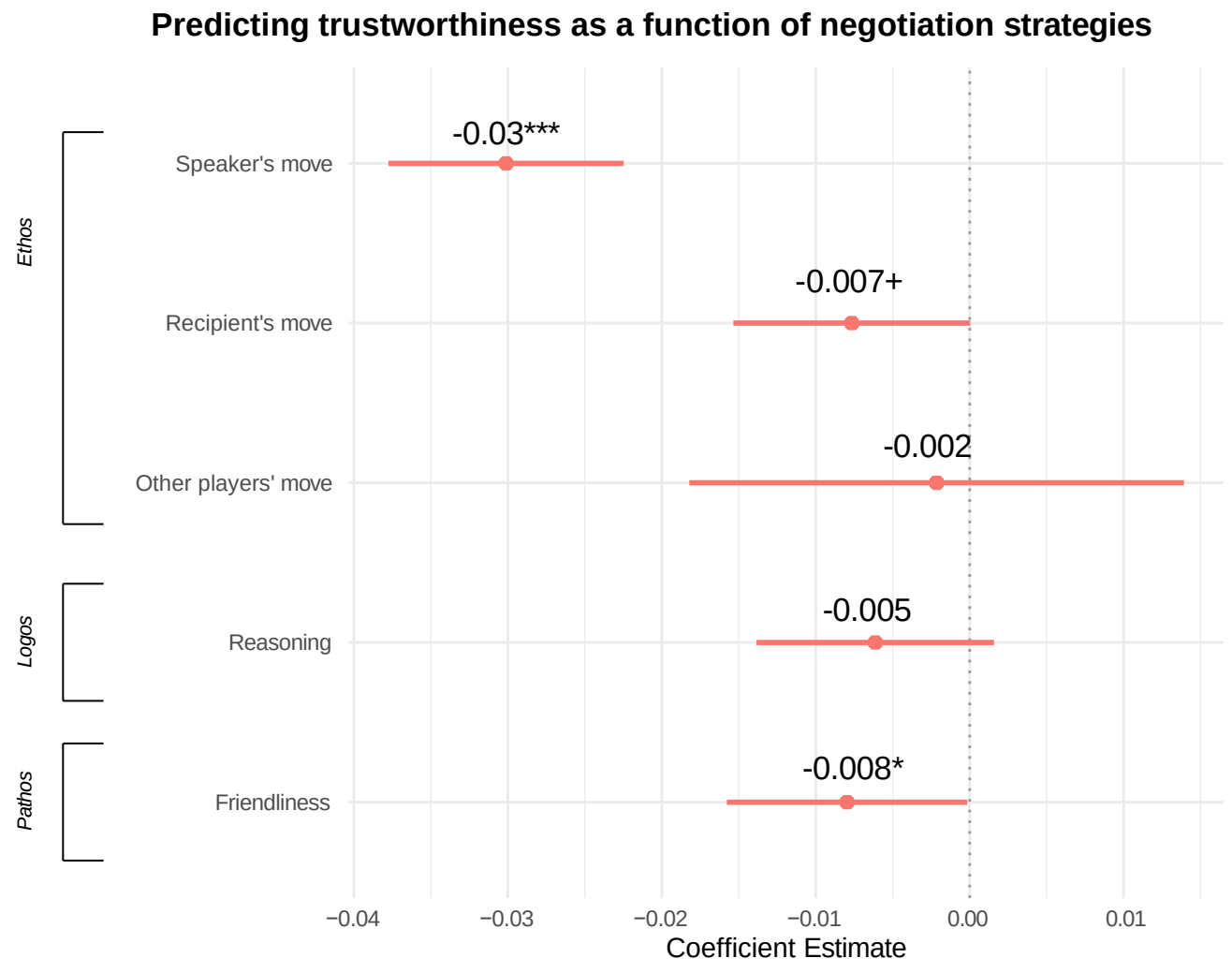

Fig. 3. Coefficient plot for predicting message-level Trustworthiness as a function of negotiation strategies. The larger the coefficient estimate, the better the negotiation strategy predicts trustworthiness. Self-disclosure (speaker's move) and friendliness are significant negative predictors of trustworthiness.

Table 4. Ablation study on Perception (of trustworthiness) prediction using individual and combinations of negotiation strategies as predictors in a DistilBERT cross-validation setup. The color gradient identifies the best performing models for each metric (darker is better). The first column reports the performance in a setup where no negotiation labels are included.

| | | | | | | | | |
|---|---|---|---|---|---|---|---|---|
| Pathos | Best | ✓ | | | ✓ | ✓ | | ✓ |
| Logos | performing | | ✓ | | ✓ | | ✓ | ✓ |
| Ethos | DistilBERT | | | ✓ | | ✓ | ✓ | ✓ |
| Accuracy | 0.842 | 0.939 | 0.939 | 0.944 | 0.937 | 0.941 | 0.943 | 0.942 |
| Macro-F1 | 0.488 | 0.505 | 0.499 | 0.499 | 0.506 | 0.508 | 0.505 | 0.512 |
| Minority-F1 | 0.062 | 0.040 | 0.031 | 0.028 | 0.045 | 0.046 | 0.038 | 0.072 |

Complementing our study of short-term consequences where we controlled for the duration of the chat conversation, here we restricted each player to their first $n$ utterances and performed the same exercise. Fig. 4b shows that the graph-aware approach outperforms others even with the first six chat messages. The vertical lines identify model performance after exchanging 25, 30, and 60 messages, which roughly correspond with the elbow point of the curves.

## 5 DISCUSSION AND IMPLICATIONS

In online games that involve collaborative play, such as Diplomacy, the dynamics of interpersonal relationships play a crucial role. Friendships can be forged, and trust can be established or broken through various actions. Communities and guilds are often formed around players who trust each other, facilitated by shared interests, values, commitment, or identity. This trust often manifests in the form of trades and gifting. However, misbehavior such as attacking, mocking, or unfulfilled trades can shatter this trust [51].

This complex interplay of trust and betrayal in online games led us to investigate the short- and long-term impact of online exchanges during Diplomacy games. Our research offers two main insights. The first key finding is the significant

| Approach | Accuracy |
|---|---|
| Random-state SBIRL | 0.530 |
| Simple SBIRL | 0.710 |
| Graph-only SBIRL | 0.640 |
| Graph-aware SBIRL | **0.790** |

(a)

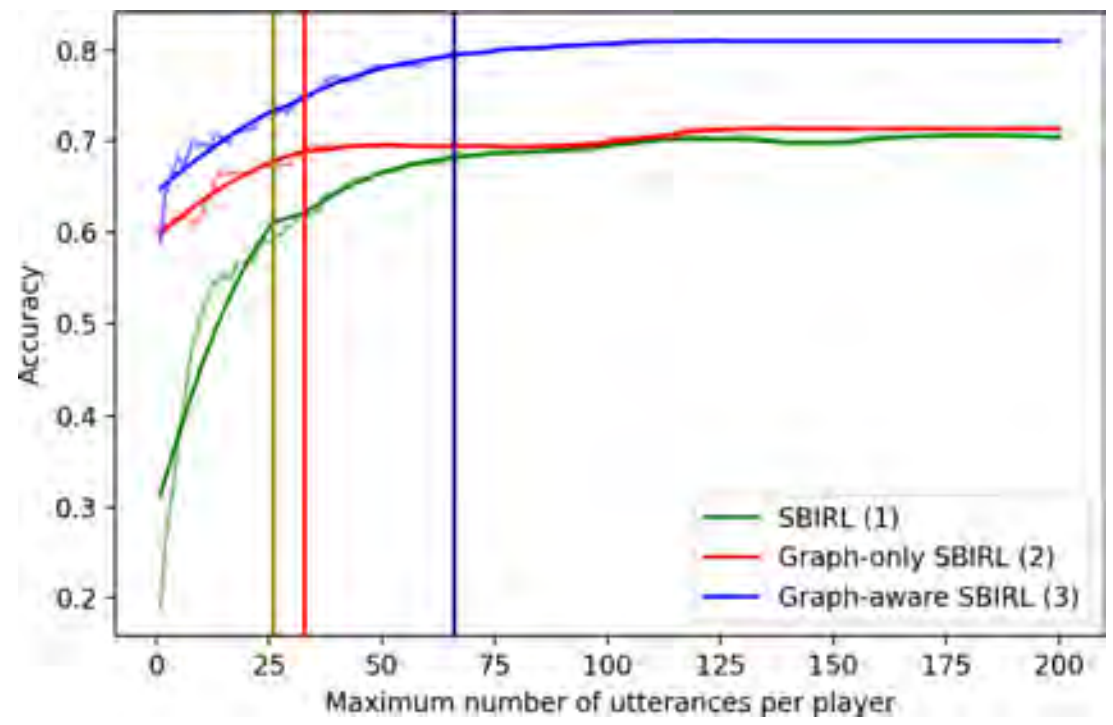

(b)

Fig. 4. (a) Accuracy of the SBIRL reward function, defined by the fraction of times the winner had a greater average estimated reward (b) Ablation analyses - winning player prediction accuracy with the number of utterances restricted. The vertical lines refer to the elbow points of the graphs, after which the accuracy plateaus.

role of various negotiation strategies in determining the perceived trustworthiness of a message. These strategies are often reflected in the linguistic choices players make during their interactions.

To evaluate the impact of these linguistic features on predicting trustworthiness, we employed machine learning models. Our analysis revealed that pronouns are key linguistic markers of a person's trustworthiness, corroborating prior work from Van Swol and Braun [71] on communicating deception through differences in language use, justifications, and questions. However, these linguistic cues can also be strategically used for deception, as suggested by Toma and Hancock [69]. For instance, Ethos involves more first-person pronouns to establish credentials and share personal information [13]. Logos involves a higher use of causal and function words to weave in asymmetric personal information for convincing the other player, as reported by Hancock et al. [33]. Lastly, Pathos involves the use of descriptive words to craft narratives, as evidenced by Birnholtz et al. [5]. These findings underscore the intricate relationship between language use, negotiation strategies, and perceived trustworthiness in online game interactions.

Our first key finding is that mentions of the speaker's move and friendliness were significant negative predictors of trustworthiness. The finding suggests that while self-disclosure and friendliness seem like rational choices for negotiation by themselves (as is apparent from the Error Analysis reported in Appendix C), when we control for other variables, such as whether or not the speaker is also mentioning others' moves, or offering logical reasoning for their speculation, their independent effect goes into the negative. Our work differs from the computational intensive work from Bakhtin et al. [3], Kramár et al. [45] that focuses on game state, and other work that focuses on Pathos – politeness, reciprocity, greetings, or friendliness for persuasion. Theoretically, we can offer a posthoc explanation that providing information about themselves or being friendly offers the least utility to the recipient in a social exchange, which is why it may be counterproductive for Diplomacy players.

Our second key finding is that incorporating the changing social dynamics into analytical models provides a distinct predictive advantage in predicting the consequences of online interactions. Prior work has not considered how dialogue affects long-term relationships or success [3]. Our finding implies that more turns of data (both textual and social network data) naturally offer more precise predictions. Additionally, we note the stability of our models to predict outcomes even 200 turns ahead. We expect this because individuals with greater centrality control the information flow between players and are therefore more likely to form a dominant coalition [7, 9, 12, 53] and are, therefore, better situated for success.

## 6 ETHICAL CONSIDERATIONS AND LIMITATIONS

This study annotated and used secondary data publicly released by previous authors with the informed consent of players participating in a game. Our work helps to develop a deeper understanding of computer-supported cooperative work, especially around persuasion, trustworthiness, and establishing a long-term rapport. However, modeling these negotiation strategies with generative models may have implications for online vulnerabilities [11]; for instance, models fine-tuned on the labeled Diplomacy dataset could work to gain someone's trust with malicious intent, including but not limited to data phishing.

Our study adheres to the FAIR princinples [75] as follows:

- **Findability**: Together with this study, we will release the annotated Diplomacy dataset and its metadata on Zenodo, a general-purpose, open-source repository developed under the OpenAIRE program managed by CERN. A unique Document Object Identifier (DOI) will thus be available. The dataset, metadata, and associated information (e.g., licenses) will be citable.
- **Accessibility**: The data and its metadata can be retrieved using standard protocols and APIs, and the metadata will remain accessible even when the dataset is no longer available.
- **Interoperability**: The data can be downloaded in JSON and exported to various formats.
- **Reusability**: We will specify the CC BY 4.0 licensed usage so that researchers may use the dataset with proper attribution.

There are a few limitations of our work. Beyond the short-term consequences of one-off exchanges, which are more susceptible to error, we recommend that scholars evaluate their models for their long-term predictive ability and stability, as much as for their precision on immediate outcomes. For such problems, models may benefit from ingesting successive data points in a temporal sequence. Our dataset comprises conversations of an online game of Diplomacy, which the players willingly share. Since Diplomacy is a niche strategy game, the participants are a sub-population of 'gamers .' The following two ethical considerations concern the replicability and generalizability of the models. First, the dataset was co-created by avid gamers familiar with the social norms of Diplomacy. Therefore, the data characteristics may be hard to replicate even when a general population of internet users is familiarized with the rules of Diplomacy and invited to play using the same experimental conditions. Likewise, the gamers' performance may be isolated in the game context and different in a real-life negotiation meeting.

A second limitation is whether the findings generalize to all internet users. Gamers are stereotyped as more introverted, prone to depression, and more socially inept than the general population [44], which would be reflected in their communication style [8]. Therefore, researchers are advised to fine-tune or domain-transfer pre-trained models to new contexts and populations. Furthermore, the data and chat message vocabulary is biased toward the gameplay mechanics. Finally, predictive models can be used to exploit online vulnerabilities [11]; for instance, pre-trained models fine-tuned on the Diplomacy dataset could work to gain someone's trust with malicious intent, including but not limited to data phishing.

## 7 CONCLUSION AND FUTURE WORK

This work offers a sociological lens to understand the nature of gameplay interactions. We identified and characterized the negotiation strategies applied in a dataset of turn-based chat messages from the online game Diplomacy. We find that a player's emerging social influence throughout a game offers the best predictions of a player's success through a study of their in-game interactions. While negotiation strategies can be predicted reasonably accurately through the textual and linguistic features of the chat messages, the negotiation strategies alone are insufficient in predicting short-term player trustworthiness. The findings signal a need to reconsider the design of studies that predict the behavioral outcomes of one-off online textual exchanges.

Our first major contribution is the negotiation taxonomy, which is essential for future work in interpersonal relationships and group dynamics, especially for persuasive contexts of organizational groups that require trust. The taxonomy could inform the design of social computing systems, online communities, and collaborative platforms, where understanding and predicting user behavior is crucial. It could also have implications for game design, particularly in games that involve social interaction and negotiation. Unlike typical MMORPGs where interactions are often driven by game mechanics, rewards, or mentorship systems, Diplomacy offers a unique environment where players rely heavily on negotiation strategies to succeed. Our study characterizes these strategies, providing insights into how players navigate this complex social landscape. For instance, our findings can inform the design of non-player characters (NPCs) in games, enabling them to use language that reflects human negotiation strategies and dynamic deception tactics, as was done in the work by Bakhtin et al. [3]. Even in multiplayer settings, NPCs that rely on a set of negotiation strategies, such as ours, can alter the story arc based on how players respond, by introducing dynamic challenges that require players to collaborate, negotiate, and strategize together, simulating real-world social interactions.

Our second major contribution is the importance of incorporating the changing social dynamics into predictive models. Our findings suggest that game design should consider the immediate gameplay mechanics and broader social dynamics that evolve over time. For instance, game designers could create mechanics that reward players for building and maintaining social connections, reflecting the benefits of social capital observed in our study.

However, it is essential to note that predicting perceived trustworthiness remains a complex task, as it may depend on various external factors that are difficult to capture in a predictive model. In future work, we aim to develop new frameworks to improve predicting the short-term outcome of exchanges with sparse data, particularly by incorporating graph and temporal features and developing and evaluating complex, non-linear parameterizations of the reward function estimator. We will also consider simulating counterfactual data to allow us to model agent behavior. Future HCI research could explore how to incorporate more contextual information into predictive models, such as user demographics, prior experience with the game, and in-game contextual information.

In future work on longer-term outcomes, we are interested in examining the effect of interplayer dynamics on group organization, evolution, and culture [34]. Finally, we are interested in examining how closely our findings can generalize to non-game setups involving social exchange, including but not limited to buyer-seller negotations [35].

**Acknowledgments:** This work was supported by a grant from the NUS Department of Communications and New Media. A special thanks to Peskov et. al [60] and Niculae et al. [52] for curating and sharing the original data. The wave 1 of annotations were also released as part of the CLAff-Diplomacy Shared Task [40]. A part of this research was published as a Student Abstract [1].

## A ANNOTATION INSTRUCTIONS

An Amazon Mechanical Turk (AMT) task was launched to obtain six annotations for each negotiation strategy in each message across the dataset. The instructions are provided in Table 1.

Table 5. A part of the instructions used on an Amazon Mechanical Turk Task to annotate the Diplomacy dataset.

These are sentences taken from people's conversations during Diplomacy games. Please classify them according to how the speaker discuss strategies and general conversation.

**Overview**

In these statements, players try to form alliances to plan military campaigns and defeat each other, but things might change quickly.

- Each statement is a piece of a dialogue from a SENDER player to a RECEIVER player.
- Please classify the statements according to whether the sender is talking about game moves, other players, reasoning out a move, or trying to build a rapport with the receiver.
- Select "YES" if you're really confident about your answer. A single statement can have a "YES" for more than one question.
- Underlined words suggest what to look out for, but there will be other signals too.

*Steps*

- Determine which categories best describe the comment.
- The statement is made by one player to another. It usually discusses the next move and why to make it. Sometimes it is simply a friendly exchange between two players.
- Review the text of the statement and help us by answering a few yes/no questions about it.
- When multiple answers are possible, select all categories that apply.

**Friendliness**

- YES: In this statement, the sender wants to be friendly with the receiver either through compliments, sharing secrets or personal thoughts, reassurances, or apologies. Examples:
  - You're my favorite.
  - Let's keep it between you and me!
  - I'm going to keep helping you as much as I can.
  - Sorry to say this.
- NO: This statement does not appear to be friendly.

## B ADDITIONAL DETAILS ABOUT THE CLASSIFICATION APPROACHES

### B.1 Linguistic feature extraction

The definitions, citations, and examples of each type of linguistic feature are reported in Table 6.

Table 6. Exemplar linguistic features used to train the machine learning classifiers. Over 42000 features were input into the feature selection and classifier training pipelines for each label.

| Feature | Definition |
|---|---|
| **Discursive features in the feature-rich classifiers** | |
| Syntactic and grammatical features | |
| **Syntactical features** | Features that are used organize information in English message, such as punctuation marks, but also social media-specific syntactical features, such as hashtags,. |
| **Grammatical features** | Parts of speech such as noun, pronoun, verb, adjective, adverb, preposition, conjunction, and interjection. They can also include linguistic units indicating the position of the feature in the text, such as text that starts with pronouns. |
| Politeness features [20] | |
| **Factuality** | Linguistic cues that are used to report a fact, e.g., "point," "reality," "truth," "actually," and "honestly." |
| **Deference** | Linguistic cues that are used to defer to another person, e.g., "great," "good," "interesting," and "awesome." |
| **Apology** | Linguistic cues that are used to issue an apology, e.g., "sorry," "forgive," and "excuse." |
| Harbingers features [52] | |
| **Claim** | Discourse connectors used to indicate a claim by the sender, e.g., "I believe," "I mean," and "I think" |
| **Premise** | Discourse connectors used to indicate assumptions and premises, e.g., "assuming that," "as indicated by," and "as shown." |
| **Comparison** | Discourse connectors used to create comparisons, e.g., "although," "by comparison," and "on the other hand." |
| Psycholinguistic features [59] | |
| **Emotional processes** | Categories used to indicate emotional expression. Individual scores for the emotional categories (anger, sadness, anxiety) are also reported. |
| **Cognitive processes** | Categories used to indicate styles of thinking and processing information. Scores are reported for markers of cognitive processing and analytical thinking, discrepancy, and comparison. |
| **Social processes** | Categories used to indicate social processes, such as mentions of other individuals and groups. |
| **Content features in the TF-IDF classifiers** | |
| **Count Vectorizer Features** | Words weighted by the frequency of their occurrence in the dataset |
| **TF-IDF features** | Words and phrases weighted by their importance in a message. The weight is measured as a ratio of the frequency of the term in the text to the frequency of the term in the overall collection of messages. |

### B.2 Classical classifiers

Classical classifiers use a set of descriptive linguistic features of the messages that are computed apriori. They then exploit the resulting feature vector as a representation of the text input for the algorithms. We used nine classification approaches that were provided by the python library *scikit-learn* [57]: K-Nearest Neighbours, Linear Support Vector Classifier, C-Support Vector Classifier, Logistic Regression, Gaussian Naive Bayes, Bernoulli Naive Bayes, Gradient Boosting, Ada Boosting.

The K-Nearest Neighbors approach classifies data points by minimizing the distances between points, and the Support Vector Classifiers minimize the distances between a point and a line. The Decision Tree approach groups data points based on similar features; the Logistic Regression approach looks for a linear relationship between the features and the label. The Naive Bayes approaches assume independence of all features and assign probabilities based on these features using a Gaussian or Bernoulli probability distribution. The boosting classifiers iteratively improve weak learners using negative gradients (Gradient Boosting) or exponential gradients (Ada Boosting) of the loss function to reduce the prediction losses [57].

### B.3 Neural network classifiers

The neural network receives the raw message as input and automatically learns the word features representing the message during the training phase. The neural network is organized in layers, in which each layer extracts some text feature before passing extracted information to more complex subsequent layers. The last layer performs a logistic regression on the representation, providing the resultant binary classification. The extracted features are specifically tailored to the classification task because they are learned via backpropagation and stochastic gradient descent, an iterative approach to optimize the loss between the predicted and actual labels.

- BERT, Bidirectional Encoder Representations from Transformers, constructs pre-trained word vectors based on both the left and right context of the target text, providing a logical and sequential connection between the texts. BERT trains the language model through masked language modeling and next sentence prediction, where it teaches the model to predict the subsequent sentence [23].
- RoBERTA evolved from BERT by removing the next sentence prediction training phase, increasing the word vocabulary range, and generating differently masked permutations for a single sentence [48].
- ALBERT made BERT more efficient through factorized embedding parameterization by breaking the word embedding matrix into smaller matrices and reusing the same parameters through the neural network layers [46].
- DistilBERT distilled, or approximated, BERT's neural network such that only half of the parameters are used while the performance remains roughly the same. It latches on the idea that once an extensive neural network is trained, a smaller network can approximate its complete output distribution [63].
- XLNet is an autoregressive model. It implements the bidirectional text contextual information by permutation of the language sequences. To do so, it predicts that the words appear in a specific position by training contextualized word vectors on random permutations of an input sequence [79].
- The Generative Pretrained Transformer 3 (GPT-3) classifier is a generative transformer classifier. It is a transformer architecture model that contains 175 billion input parameters and has been pre-trained on an extraordinarily vast amount of text data, which includes the Common Crawl, WebText2, Books1, Books2, and Wikipedia Corpora. Its predictive accuracy is anticipated to be the best among the neural network classifiers due to its huge database and input parameters [10].

## C ADDITIONAL RESULTS

Detailed results about evaluating different classical classifiers trained on different sets of linguistic features, including the ones finally chosen to train the models for each label (best Minority-F1 score), are provided in Tables 7-10. Table 11 reports the neural network classifiers.

From Table 10, we observe that classifiers trained on a combination of TFIDF and discursive features had an advantage over solely TFIDF features. Furthermore, the logistic regression classifiers usually performed the best. We further observe a relatively high macro-F1 score with a low standard deviation (mean = 0.66, standard deviation = 0.094). The macro F1 scores show that the classifiers trained on Recipient's Move performed best (macro F1 = 0.741).

Table 7. All results from the internal validation of the **discursive-feature** based machine learning classifiers on the held-out test set. A score closer to 1 implies that a greater number of cases were correctly predicted as positive or negative.

| Approach | 1<br>Accuracy | 2<br>Macro F1 | 3<br>Minority F1 | 4<br>Recall | 5<br>Precision |
|---|---|---|---|---|---|
| | **Other Player's Move** | | | | |
| Logistic regression | 0.677 | 0.550 | 0.311 | 0.758 | 0.196 |
| K-Nearest neighbors | 0.898 | 0.509 | 0.071 | 0.041 | 0.301 |
| Gaussian naive bayes | 0.339 | 0.318 | 0.203 | 0.878 | 0.115 |
| Bernoulli naive bayes | 0.758 | 0.547 | 0.239 | 0.395 | 0.171 |
| Adaboost | 0.906 | 0.621 | 0.292 | 0.204 | 0.519 |
| Gradient boosting | 0.912 | 0.613 | 0.273 | 0.175 | 0.661 |
| Decision tree | 0.863 | 0.605 | 0.286 | 0.286 | 0.288 |
| Linear support vector | 0.747 | 0.590 | 0.336 | 0.663 | 0.226 |
| C-support vector | 0.819 | 0.625 | 0.356 | 0.506 | 0.277 |
| | **Speaker's Move** | | | | |
| Logistic regression | 0.706 | 0.666 | 0.552 | 0.710 | 0.452 |
| K-Nearest neighbors | 0.703 | 0.586 | 0.366 | 0.337 | 0.401 |
| Gaussian naive bayes | 0.516 | 0.509 | 0.462 | 0.816 | 0.323 |
| Bernoulli naive bayes | 0.713 | 0.639 | 0.475 | 0.510 | 0.444 |
| Adaboost | 0.771 | 0.652 | 0.448 | 0.367 | 0.577 |
| Gradient boosting | 0.777 | 0.642 | 0.423 | 0.323 | 0.618 |
| Decision tree | 0.711 | 0.612 | 0.416 | 0.407 | 0.428 |
| Linear support vector | 0.707 | 0.666 | 0.550 | 0.704 | 0.452 |
| C-support vector | 0.723 | 0.671 | 0.541 | 0.642 | 0.469 |

| Approach | 1<br>Accuracy | 2<br>Macro F1 | 3<br>Minority F1 | 4<br>Recall | 5<br>Precision |
|---|---|---|---|---|---|
| | **Recipient's Move** | | | | |
| Logistic regression | 0.709 | 0.707 | 0.681 | 0.693 | 0.669 |
| K-Nearest neighbors | 0.623 | 0.623 | 0.621 | 0.693 | 0.564 |
| Gaussian naive bayes | 0.607 | 0.600 | 0.648 | 0.812 | 0.540 |
| Bernoulli naive bayes | 0.682 | 0.677 | 0.637 | 0.626 | 0.650 |
| Adaboost | 0.714 | 0.705 | 0.653 | 0.604 | 0.713 |
| Gradient boosting | 0.723 | 0.713 | 0.660 | 0.603 | 0.731 |
| Decision tree | 0.640 | 0.634 | 0.587 | 0.574 | 0.602 |
| Linear support vector | 0.704 | 0.702 | 0.680 | 0.707 | 0.657 |
| C-support vector | 0.705 | 0.700 | 0.662 | 0.646 | 0.679 |
| | **Reasoning** | | | | |
| Logistic regression | 0.564 | 0.507 | 0.341 | 0.570 | 0.826 |
| K-Nearest neighbors | 0.762 | 0.479 | 0.095 | 0.946 | 0.793 |
| Gaussian naive bayes | 0.630 | 0.533 | 0.319 | 0.686 | 0.818 |
| Bernoulli naive bayes | 0.730 | 0.541 | 0.246 | 0.866 | 0.807 |
| Adaboost | 0.788 | 0.452 | 0.023 | 0.992 | 0.793 |
| Gradient boosting | 0.790 | 0.445 | 0.007 | 0.996 | 0.792 |
| Decision tree | 0.665 | 0.504 | 0.220 | 0.780 | 0.794 |
| Linear support vector | 0.600 | 0.521 | 0.327 | 0.634 | 0.820 |
| C-support vector | 0.653 | 0.531 | 0.292 | 0.733 | 0.810 |
| | **Friendliness** | | | | |
| Logistic regression | 0.673 | 0.659 | 0.593 | 0.663 | 0.803 |
| K-Nearest neighbors | 0.628 | 0.588 | 0.461 | 0.714 | 0.716 |
| Gaussian naive bayes | 0.477 | 0.465 | 0.543 | 0.253 | 0.830 |
| Bernoulli naive bayes | 0.638 | 0.611 | 0.510 | 0.687 | 0.741 |
| Adaboost | 0.706 | 0.635 | 0.475 | 0.874 | 0.730 |
| Gradient boosting | 0.713 | 0.642 | 0.482 | 0.884 | 0.733 |
| Decision tree | 0.653 | 0.613 | 0.490 | 0.742 | 0.732 |
| Linear support vector | 0.678 | 0.663 | 0.593 | 0.677 | 0.801 |
| C-support vector | 0.699 | 0.671 | 0.574 | 0.757 | 0.778 |

### C.1 Final prediction of Trustworthiness

The predictive performance of using the best-performing classifier set up for Perception on all the labels is reported in Table 12. It reports the model performance for the combined classifiers, including negotiation strategy labels as inputs. However, we observe that the poor Minority-F1 score implies difficulty predicting trustworthiness with linguistic features alone.

Examining a confusion matrix of the final label of Perception, reported in Figure 5, we observe that Lies have the highest true positive rate, indicating that the linguistic features used in constructing a Lie are very distinct. On the other hand, machine classifiers have difficulty identifying Truth statements and thus struggle with predicting trustworthiness.

*C.1.1 Error Analysis.* Classical classifiers trained on the Trustworthiness label with linguistic features pay attention to references to the self (*I, my*) which have a negative feature attribution for the Trustworthiness label, and to the 'friend' category from LIWC (*friend, buddy*), which have a positive feature attribution for the Trustworthiness label, as seen in Figure 6.

Table 8. All results from the internal validation of the **word-feature** based machine learning classifiers on the held-out test set. A score closer to 1 implies that a greater number of cases were correctly predicted as positive or negative.

| Approach | 1<br>Accuracy | 2<br>Macro F1 | 3<br>Minority F1 | 4<br>Recall | 5<br>Precision |
|---|---|---|---|---|---|
| **Other Player's Move** | | | | | |
| Logistic regression | 0.833 | 0.634 | 0.363 | 0.497 | 0.288 |
| K-Nearest neighbors | 0.903 | 0.507 | 0.065 | 0.035 | 0.388 |
| Gaussian naive bayes | 0.603 | 0.448 | 0.157 | 0.390 | 0.098 |
| Bernoulli naive bayes | 0.902 | 0.494 | 0.039 | 0.021 | 0.311 |
| Adaboost | 0.900 | 0.565 | 0.182 | 0.118 | 0.424 |
| Gradient boosting | 0.905 | 0.512 | 0.074 | 0.040 | 0.540 |
| Decision tree | 0.869 | 0.552 | 0.175 | 0.146 | 0.222 |
| Linear support vector | 0.834 | 0.604 | 0.303 | 0.378 | 0.254 |
| C-support vector | 0.878 | 0.638 | 0.343 | 0.333 | 0.359 |
| **Speaker's Move** | | | | | |
| Logistic regression | 0.682 | 0.612 | 0.447 | 0.506 | 0.401 |
| K-Nearest neighbors | 0.745 | 0.451 | 0.050 | 0.027 | 0.466 |
| Gaussian naive bayes | 0.407 | 0.404 | 0.361 | 0.660 | 0.248 |
| Bernoulli naive bayes | 0.741 | 0.534 | 0.223 | 0.147 | 0.471 |
| Adaboost | 0.739 | 0.529 | 0.214 | 0.141 | 0.461 |
| Gradient boosting | 0.747 | 0.475 | 0.098 | 0.055 | 0.519 |
| Decision tree | 0.696 | 0.558 | 0.312 | 0.273 | 0.366 |
| Linear support vector | 0.680 | 0.600 | 0.422 | 0.460 | 0.390 |
| C-support vector | 0.706 | 0.624 | 0.447 | 0.467 | 0.430 |

| Approach | 1<br>Accuracy | 2<br>Macro F1 | 3<br>Minority F1 | 4<br>Recall | 5<br>Precision |
|---|---|---|---|---|---|
| **Recipient's Move** | | | | | |
| Logistic regression | 0.715 | 0.710 | 0.671 | 0.650 | 0.694 |
| K-Nearest neighbors | 0.662 | 0.619 | 0.492 | 0.366 | 0.750 |
| Gaussian naive bayes | 0.513 | 0.480 | 0.611 | 0.855 | 0.475 |
| Bernoulli naive bayes | 0.719 | 0.711 | 0.663 | 0.620 | 0.713 |
| Adaboost | 0.706 | 0.692 | 0.626 | 0.552 | 0.725 |
| Gradient boosting | 0.705 | 0.680 | 0.590 | 0.476 | 0.778 |
| Decision tree | 0.664 | 0.656 | 0.605 | 0.577 | 0.638 |
| Linear support vector | 0.682 | 0.678 | 0.642 | 0.639 | 0.646 |
| C-support vector | 0.725 | 0.718 | 0.674 | 0.638 | 0.716 |
| **Reasoning** | | | | | |
| Logistic regression | 0.595 | 0.501 | 0.286 | 0.649 | 0.802 |
| K-Nearest neighbors | 0.755 | 0.480 | 0.101 | 0.936 | 0.793 |
| Gaussian naive bayes | 0.406 | 0.398 | 0.328 | 0.330 | 0.807 |
| Bernoulli naive bayes | 0.779 | 0.460 | 0.044 | 0.977 | 0.792 |
| Adaboost | 0.789 | 0.445 | 0.008 | 0.995 | 0.792 |
| Gradient boosting | 0.792 | 0.444 | 0.005 | 0.999 | 0.792 |
| Decision tree | 0.699 | 0.502 | 0.189 | 0.838 | 0.793 |
| Linear support vector | 0.603 | 0.504 | 0.282 | 0.663 | 0.802 |
| C-support vector | 0.621 | 0.521 | 0.303 | 0.680 | 0.811 |
| **Friendliness** | | | | | |
| Logistic regression | 0.745 | 0.724 | 0.650 | 0.776 | 0.824 |
| K-Nearest neighbors | 0.701 | 0.612 | 0.426 | 0.900 | 0.716 |
| Gaussian naive bayes | 0.496 | 0.495 | 0.514 | 0.350 | 0.745 |
| Bernoulli naive bayes | 0.723 | 0.671 | 0.540 | 0.856 | 0.754 |
| Adaboost | 0.724 | 0.670 | 0.537 | 0.860 | 0.754 |
| Gradient boosting | 0.721 | 0.656 | 0.505 | 0.884 | 0.741 |
| Decision tree | 0.690 | 0.659 | 0.557 | 0.756 | 0.768 |
| Linear support vector | 0.714 | 0.690 | 0.606 | 0.753 | 0.798 |
| C-support vector | 0.744 | 0.727 | 0.658 | 0.761 | 0.835 |

Table 9. All results from the internal validation of the **tfidf-features** based machine learning classifiers on the held-out test set. A score closer to 1 implies that a greater number of cases were correctly predicted as positive or negative.

| Approach | 1<br>Accuracy | 2<br>Macro F1 | 3<br>Minority F1 | 4<br>Recall | 5<br>Precision |
|---|---|---|---|---|---|
| **Other Player's Move** | | | | | |
| Logistic regression | 0.814 | 0.632 | 0.374 | 0.580 | 0.277 |
| K-Nearest neighbors | 0.900 | 0.486 | 0.025 | 0.014 | 0.162 |
| Gaussian naive bayes | 0.603 | 0.448 | 0.157 | 0.390 | 0.098 |
| Bernoulli naive bayes | 0.902 | 0.494 | 0.039 | 0.021 | 0.311 |
| Adaboost | 0.898 | 0.576 | 0.206 | 0.139 | 0.408 |
| Gradient boosting | 0.903 | 0.506 | 0.063 | 0.035 | 0.417 |
| Decision tree | 0.861 | 0.564 | 0.203 | 0.185 | 0.225 |
| Linear support vector | 0.832 | 0.618 | 0.331 | 0.436 | 0.268 |
| C-support vector | 0.899 | 0.617 | 0.289 | 0.219 | 0.439 |
| **Speaker's Move** | | | | | |
| Logistic regression | 0.669 | 0.614 | 0.468 | 0.572 | 0.396 |
| K-Nearest neighbors | 0.732 | 0.468 | 0.092 | 0.054 | 0.351 |
| Gaussian naive bayes | 0.408 | 0.405 | 0.360 | 0.656 | 0.248 |
| Bernoulli naive bayes | 0.741 | 0.534 | 0.223 | 0.147 | 0.471 |
| Adaboost | 0.738 | 0.537 | 0.232 | 0.158 | 0.449 |
| Gradient boosting | 0.746 | 0.475 | 0.098 | 0.055 | 0.509 |
| Decision tree | 0.685 | 0.562 | 0.330 | 0.306 | 0.360 |
| Linear support vector | 0.668 | 0.601 | 0.438 | 0.509 | 0.385 |
| C-support vector | 0.709 | 0.612 | 0.418 | 0.412 | 0.425 |

| Approach | 1<br>Accuracy | 2<br>Macro F1 | 3<br>Minority F1 | 4<br>Recall | 5<br>Precision |
|---|---|---|---|---|---|
| **Recipient's Move** | | | | | |
| Logistic regression | 0.719 | 0.715 | 0.679 | 0.665 | 0.694 |
| K-Nearest neighbors | 0.631 | 0.601 | 0.492 | 0.402 | 0.645 |
| Gaussian naive bayes | 0.515 | 0.485 | 0.608 | 0.843 | 0.476 |
| Bernoulli naive bayes | 0.719 | 0.711 | 0.663 | 0.620 | 0.713 |
| Adaboost | 0.706 | 0.693 | 0.629 | 0.558 | 0.722 |
| Gradient boosting | 0.702 | 0.676 | 0.586 | 0.474 | 0.771 |
| Decision tree | 0.651 | 0.644 | 0.597 | 0.580 | 0.616 |
| Linear support vector | 0.688 | 0.685 | 0.655 | 0.664 | 0.647 |
| C-support vector | 0.716 | 0.709 | 0.664 | 0.629 | 0.705 |
| **Reasoning** | | | | | |
| Logistic regression | 0.596 | 0.503 | 0.288 | 0.650 | 0.803 |
| K-Nearest neighbors | 0.767 | 0.469 | 0.072 | 0.957 | 0.792 |
| Gaussian naive bayes | 0.407 | 0.398 | 0.328 | 0.330 | 0.807 |
| Bernoulli naive bayes | 0.779 | 0.460 | 0.044 | 0.977 | 0.792 |
| Adaboost | 0.784 | 0.455 | 0.032 | 0.985 | 0.793 |
| Gradient boosting | 0.791 | 0.443 | 0.003 | 0.999 | 0.792 |
| Decision tree | 0.689 | 0.506 | 0.206 | 0.820 | 0.795 |
| Linear support vector | 0.610 | 0.506 | 0.279 | 0.675 | 0.801 |
| C-support vector | 0.712 | 0.514 | 0.203 | 0.853 | 0.798 |
| **Friendliness** | | | | | |
| Logistic regression | 0.748 | 0.731 | 0.663 | 0.764 | 0.836 |
| K-Nearest neighbors | 0.690 | 0.607 | 0.426 | 0.878 | 0.715 |
| Gaussian naive bayes | 0.496 | 0.495 | 0.512 | 0.353 | 0.742 |
| Bernoulli naive bayes | 0.723 | 0.671 | 0.540 | 0.856 | 0.754 |
| Adaboost | 0.724 | 0.672 | 0.541 | 0.856 | 0.755 |
| Gradient boosting | 0.724 | 0.656 | 0.504 | 0.890 | 0.740 |
| Decision tree | 0.687 | 0.655 | 0.550 | 0.756 | 0.763 |
| Linear support vector | 0.729 | 0.709 | 0.635 | 0.753 | 0.818 |
| C-support vector | 0.748 | 0.727 | 0.653 | 0.780 | 0.825 |

We see differences in word attributions by the best-performing neural network models in Figure 7, as the model appears to positively weigh self-references, indicating that the speaker is discussing their move. However, this appears to lead to poor performance, where the models overfit to the positive class and report a poor minority-F1 score.

Table 10. All results from the internal validation of the **tfidf-discursive-features** based machine learning classifiers on the held-out test set. A score closer to 1 implies that a greater number of cases were correctly predicted as positive or negative.

| Approach | 1<br>Accuracy | 2<br>Macro F1 | 3<br>Minority F1 | 4<br>Recall | 5<br>Precision |
|---|---|---|---|---|---|
| | **Other Player's Move** | | | | |
| Logistic regression | 0.838 | 0.675 | 0.445 | 0.675 | 0.333 |
| K-Nearest neighbors | 0.898 | 0.509 | 0.071 | 0.041 | 0.301 |
| Gaussian naive bayes | 0.603 | 0.448 | 0.156 | 0.386 | 0.098 |
| Bernoulli naive bayes | 0.898 | 0.524 | 0.102 | 0.062 | 0.315 |
| Adaboost | 0.908 | 0.654 | 0.358 | 0.270 | 0.548 |
| Gradient boosting | 0.911 | 0.590 | 0.227 | 0.136 | 0.693 |
| Decision tree | 0.868 | 0.602 | 0.277 | 0.264 | 0.293 |
| Linear support vector | 0.860 | 0.664 | 0.407 | 0.503 | 0.343 |
| C-support vector | 0.832 | 0.638 | 0.373 | 0.508 | 0.298 |
| | **Speaker's Move** | | | | |
| Logistic regression | 0.731 | 0.685 | 0.566 | 0.689 | 0.480 |
| K-Nearest neighbors | 0.704 | 0.588 | 0.369 | 0.341 | 0.404 |
| Gaussian naive bayes | 0.408 | 0.405 | 0.360 | 0.654 | 0.248 |
| Bernoulli naive bayes | 0.752 | 0.637 | 0.432 | 0.370 | 0.520 |
| Adaboost | 0.773 | 0.662 | 0.467 | 0.392 | 0.580 |
| Gradient boosting | 0.778 | 0.631 | 0.398 | 0.290 | 0.640 |
| Decision tree | 0.717 | 0.618 | 0.423 | 0.409 | 0.441 |
| Linear support vector | 0.736 | 0.676 | 0.537 | 0.603 | 0.485 |
| C-support vector | 0.725 | 0.674 | 0.544 | 0.645 | 0.472 |

| Approach | 1<br>Accuracy | 2<br>Macro F1 | 3<br>Minority F1 | 4<br>Recall | 5<br>Precision |
|---|---|---|---|---|---|
| | **Recipient's Move** | | | | |
| Logistic regression | 0.743 | 0.741 | 0.717 | 0.727 | 0.707 |
| K-Nearest neighbors | 0.627 | 0.626 | 0.626 | 0.701 | 0.567 |
| Gaussian naive bayes | 0.520 | 0.493 | 0.610 | 0.839 | 0.479 |
| Bernoulli naive bayes | 0.728 | 0.722 | 0.680 | 0.646 | 0.717 |
| Adaboost | 0.733 | 0.726 | 0.681 | 0.640 | 0.729 |
| Gradient boosting | 0.746 | 0.738 | 0.692 | 0.638 | 0.755 |
| Decision tree | 0.681 | 0.676 | 0.637 | 0.628 | 0.648 |
| Linear support vector | 0.718 | 0.716 | 0.689 | 0.698 | 0.681 |
| C-support vector | 0.705 | 0.700 | 0.661 | 0.643 | 0.680 |
| | **Reasoning** | | | | |
| Logistic regression | 0.601 | 0.515 | 0.311 | 0.645 | 0.813 |
| K-Nearest neighbors | 0.763 | 0.480 | 0.097 | 0.946 | 0.794 |
| Gaussian naive bayes | 0.407 | 0.398 | 0.328 | 0.330 | 0.807 |
| Bernoulli naive bayes | 0.771 | 0.490 | 0.111 | 0.955 | 0.796 |
| Adaboost | 0.785 | 0.457 | 0.034 | 0.986 | 0.793 |
| Gradient boosting | 0.792 | 0.444 | 0.005 | 0.999 | 0.792 |
| Decision tree | 0.685 | 0.504 | 0.206 | 0.813 | 0.794 |
| Linear support vector | 0.627 | 0.518 | 0.288 | 0.696 | 0.807 |
| C-support vector | 0.656 | 0.533 | 0.294 | 0.737 | 0.811 |
| | **Friendliness** | | | | |
| Logistic regression | 0.740 | 0.724 | 0.656 | 0.751 | 0.835 |
| K-Nearest neighbors | 0.626 | 0.589 | 0.464 | 0.709 | 0.717 |
| Gaussian naive bayes | 0.496 | 0.496 | 0.512 | 0.355 | 0.742 |
| Bernoulli naive bayes | 0.732 | 0.699 | 0.599 | 0.811 | 0.786 |
| Adaboost | 0.729 | 0.681 | 0.558 | 0.852 | 0.762 |
| Gradient boosting | 0.725 | 0.667 | 0.528 | 0.872 | 0.749 |
| Decision tree | 0.682 | 0.650 | 0.545 | 0.751 | 0.761 |
| Linear support vector | 0.725 | 0.705 | 0.628 | 0.752 | 0.814 |
| C-support vector | 0.702 | 0.674 | 0.579 | 0.759 | 0.780 |

Table 11. All results from the internal validation of the neural network machine learning classifiers on the held-out test set. A score closer to 1 implies that a greater number of cases were correctly predicted as positive or negative.

| Approach | 1<br>Accuracy | 2<br>Macro F1 | 3<br>Minority F1 | 4<br>Recall | 5<br>Precision |
|---|---|---|---|---|---|
| | **Recipient's Move** | | | | |
| BERT | 0.773 | 0.770 | 0.742 | 0.729 | 0.756 |
| **RoBERTa** | 0.801 | 0.799 | 0.781 | 0.795 | 0.769 |
| ALBERT | 0.784 | 0.782 | 0.760 | 0.765 | 0.755 |
| DistilBERT | 0.769 | 0.767 | 0.743 | 0.744 | 0.742 |
| XLNet | 0.798 | 0.796 | 0.780 | 0.801 | 0.760 |
| GPT-3 | 0.600 | 0.575 | 0.474 | 0.405 | 0.572 |
| | **Other Player's Move** | | | | |
| BERT | | 0.724 | 0.494 | | |
| RoBERTa | 0.927 | 0.686 | 0.411 | 0.400 | 0.425 |
| **ALBERT** | 0.933 | 0.803 | 0.644 | 0.631 | 0.662 |
| DistilBERT | 0.910 | 0.720 | 0.489 | 0.452 | 0.536 |
| XLNet | 0.934 | 0.771 | 0.578 | 0.540 | 0.630 |
| GPT-3 | 0.870 | 0.543 | 0.156 | 0.126 | 0.208 |
| | **Speaker's Move** | | | | |
| BERT | | | 0.600 | | |
| **RoBERTa** | 0.834 | 0.783 | 0.678 | 0.690 | 0.667 |
| ALBERT | 0.820 | 0.761 | 0.643 | 0.636 | 0.651 |
| DistilBERT | 0.795 | 0.722 | 0.580 | 0.557 | 0.606 |
| XLNet | 0.827 | 0.774 | 0.665 | 0.677 | 0.656 |
| GPT-3 | 0.694 | 0.530 | 0.252 | 0.203 | 0.332 |

| Approach | 1<br>Accuracy | 2<br>Macro F1 | 3<br>Minority F1 | 4<br>Recall | 5<br>Precision |
|---|---|---|---|---|---|
| | **Reasoning** | | | | |
| BERT | 0.734 | 0.492 | 0.143 | 0.893 | 0.796 |
| RoBERTa | 0.775 | 0.460 | 0.050 | 0.965 | 0.795 |
| ALBERT | 0.744 | 0.472 | 0.101 | 0.904 | 0.802 |
| **DistilBERT** | 0.712 | 0.514 | 0.204 | 0.852 | 0.799 |
| XLNet | 0.563 | 0.381 | 0.192 | 0.599 | 0.562 |
| GPT-3 | 0.570 | 0.479 | 0.261 | 0.624 | 0.789 |
| | **Friendliness** | | | | |
| BERT | 0.739 | 0.708 | 0.614 | 0.813 | 0.794 |
| RoBERTa | 0.747 | 0.679 | 0.544 | 0.843 | 0.802 |
| **ALBERT** | 0.760 | 0.738 | 0.663 | 0.798 | 0.829 |
| DistilBERT | 0.733 | 0.705 | 0.614 | 0.795 | 0.797 |
| XLNet | 0.761 | 0.737 | 0.658 | 0.810 | 0.822 |
| GPT-3 | 0.542 | 0.538 | 0.497 | 0.482 | 0.727 |

Table 12. All results from the combined machine learning classifiers on the held-out test set. The classifier uses the facets Game Move, Other Player's Move, Speaker's Move, Recipient's Move, Reasoning and Friendliness to predict the perception label. A score closer to 1 implies that a greater number of cases were correctly predicted as positive or negative.

| Approach | 1<br>Accuracy | 2<br>Macro F1 | 3<br>Minority F1 | 4<br>Recall | 5<br>Precision |
|---|---|---|---|---|---|
| | **discursive-features** | | | | |
| Logistic regression | 0.595 | 0.418 | 0.098 | 0.426 | 0.055 |
| K-Nearest neighbors | 0.948 | 0.490 | 0.006 | 0.003 | 0.100 |
| Gaussian naive bayes | 0.146 | 0.143 | 0.098 | 0.901 | 0.052 |
| Bernoulli naive bayes | 0.942 | 0.504 | 0.038 | 0.023 | 0.148 |
| Adaboost | 0.948 | 0.487 | 0.000 | 0.000 | 0.000 |
| Gradient boosting | 0.946 | 0.486 | 0.000 | 0.000 | 0.000 |
| Decision tree | 0.886 | 0.491 | 0.043 | 0.049 | 0.038 |
| Linear support vector | 0.790 | 0.485 | 0.089 | 0.198 | 0.058 |
| C-support vector | 0.885 | 0.509 | 0.080 | 0.099 | 0.068 |
| | **word-features** | | | | |
| Logistic regression | | | 0.108 | | |
| K-Nearest neighbors | 0.947 | 0.490 | 0.006 | 0.003 | 0.100 |
| Gaussian naive bayes | 0.754 | 0.472 | 0.087 | 0.231 | 0.054 |
| Bernoulli naive bayes | 0.948 | 0.490 | 0.006 | 0.003 | 0.100 |
| Adaboost | 0.947 | 0.493 | 0.013 | 0.007 | 0.150 |
| Gradient boosting | 0.947 | 0.490 | 0.006 | 0.003 | 0.100 |
| Decision tree | 0.920 | 0.501 | 0.043 | 0.036 | 0.054 |
| Linear support vector | 0.851 | 0.501 | 0.084 | 0.132 | 0.061 |
| C-support vector | 0.936 | 0.493 | 0.019 | 0.013 | 0.040 |
| | **tfidf-features** | | | | |
| Logistic regression | | | 0.114 | | |
| K-Nearest neighbors | 0.947 | 0.486 | 0.000 | 0.000 | 0.000 |
| Gaussian naive bayes | 0.754 | 0.472 | 0.087 | 0.231 | 0.054 |
| Bernoulli naive bayes | 0.948 | 0.490 | 0.006 | 0.003 | 0.100 |
| Adaboost | 0.944 | 0.492 | 0.012 | 0.007 | 0.070 |
| Gradient boosting | 0.948 | 0.490 | 0.006 | 0.003 | 0.100 |
| Decision tree | 0.908 | 0.510 | 0.068 | 0.066 | 0.072 |
| Linear support vector | 0.852 | 0.505 | 0.091 | 0.145 | 0.067 |
| C-support vector | 0.930 | 0.493 | 0.023 | 0.016 | 0.043 |

| Approach | 1<br>Accuracy | 2<br>Macro F1 | 3<br>Minority F1 | 4<br>Recall | 5<br>Precision |
|---|---|---|---|---|---|
| | **tfidf-discursive-features** | | | | |
| Logistic regression | 0.811 | 0.501 | 0.107 | 0.221 | 0.071 |
| K-Nearest neighbors | 0.948 | 0.490 | 0.006 | 0.003 | 0.100 |
| Gaussian naive bayes | 0.755 | 0.472 | 0.085 | 0.224 | 0.053 |
| Bernoulli naive bayes | 0.947 | 0.490 | 0.006 | 0.003 | 0.050 |
| Adaboost | 0.943 | 0.488 | 0.006 | 0.003 | 0.033 |
| Gradient boosting | 0.948 | 0.487 | 0.000 | 0.000 | 0.000 |
| Decision tree | 0.907 | 0.514 | 0.078 | 0.076 | 0.082 |
| Linear support vector | 0.867 | 0.509 | 0.090 | 0.129 | 0.069 |
| C-support vector | 0.893 | 0.510 | 0.076 | 0.086 | 0.069 |
| | **neural network classifiers** | | | | |
| BERT | 0.936 | 0.501 | 0.036 | 0.026 | 0.065 |
| RoBERTa | 0.946 | 0.489 | 0.005 | 0.003 | 0.010 |
| ALBERT | 0.949 | 0.487 | 0.000 | 0.000 | 0.000 |
| DistilBERT | 0.929 | 0.518 | 0.072 | 0.056 | 0.111 |
| XLNet | 0.944 | 0.497 | 0.023 | 0.017 | 0.046 |
| GPT-3 | 0.925 | 0.512 | 0.063 | 0.050 | 0.088 |

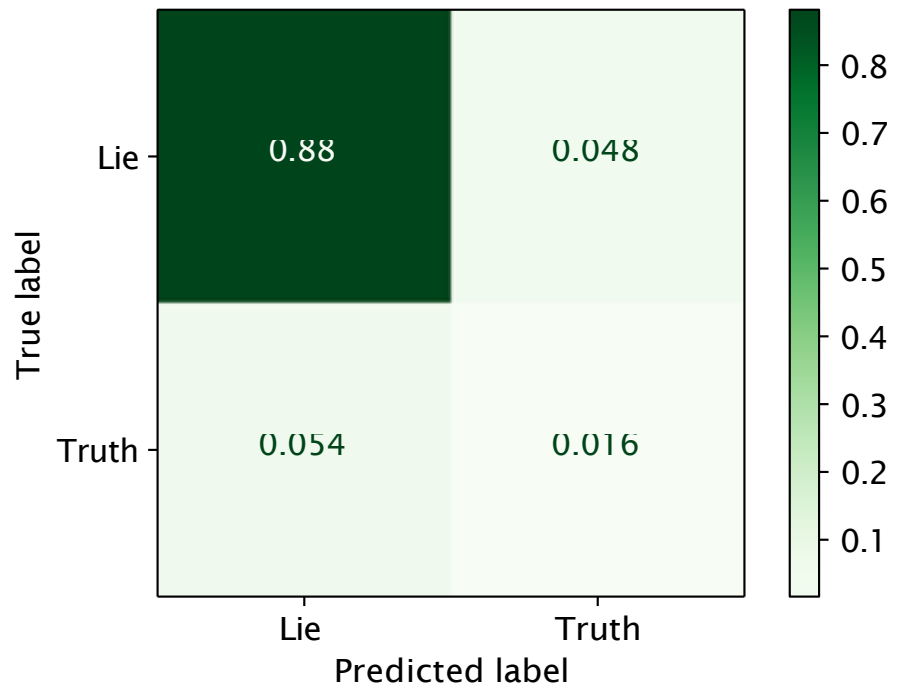

Fig. 5. Confusion Matrix of prediction of Perception

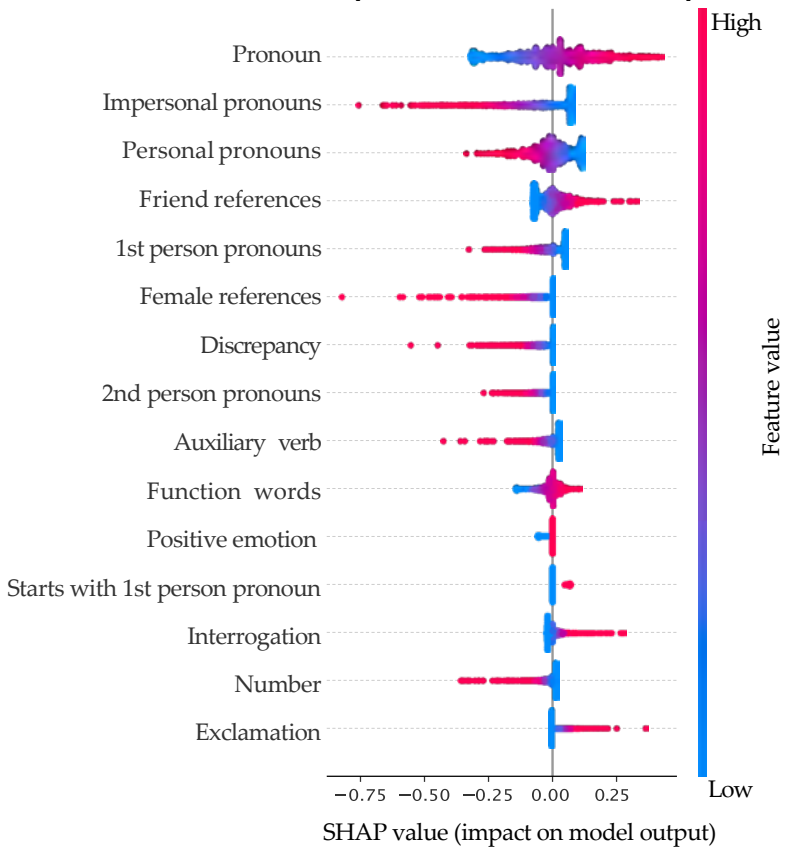

Fig. 6. Shapley value plots denoting the importance of different linguistic features in the classical classifiers. Best seen in color.

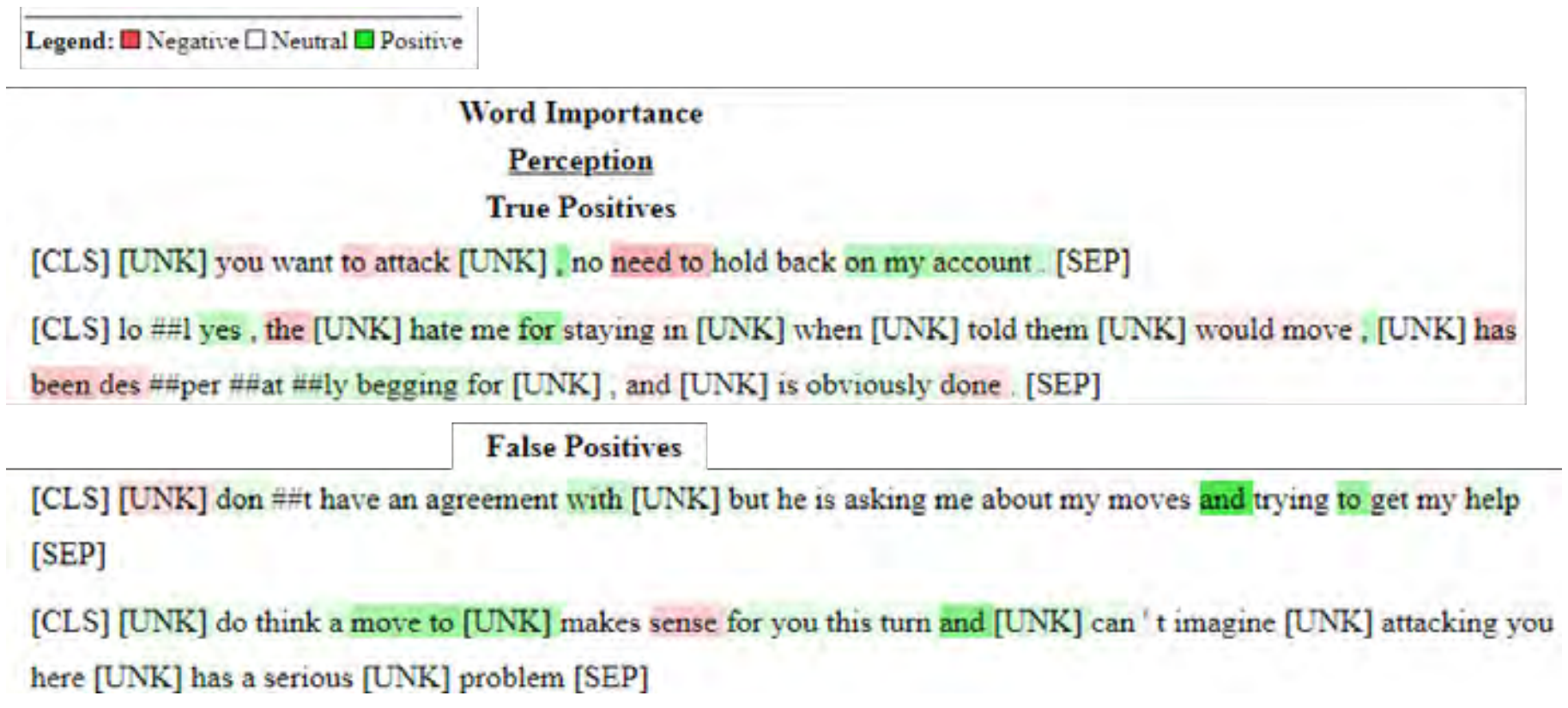

Fig. 7. Word attributions for positive classification for the final label Trustworthiness. Words highlighted in green (red) positively (negatively) attributed to the outcome; the darker the highlight, the higher the attribution. Best seen in color.